\documentclass[10pt,twocolumn,letterpaper]{article}

\usepackage{iccv}              

\definecolor{iccvblue}{rgb}{0.21,0.49,0.74}
\usepackage[pagebackref,breaklinks,colorlinks,citecolor=iccvblue,linkcolor=red,anchorcolor=magenta]{hyperref}

\usepackage[accsupp]{axessibility} 

\usepackage{algorithm}
\usepackage{algpseudocode}

\usepackage{multirow}
\usepackage{adjustbox}
\usepackage{arydshln}

\usepackage{fix-cm}

\def\paperID{8152} 
\def\confName{ICCV}
\def\confYear{2025}

\title{PanoLlama: Generating Endless and Coherent Panoramas with Next-Token-Prediction LLMs}

\author{Teng Zhou\(^{1}\), Xiaoyu Zhang\(^{1}\), Yongchuan Tang\(^{1}\)\thanks{Corresponding Author} \\
\(^{1}\)College of Computer Science and Technology, Zhejiang University \\
}

\begin{document}
\maketitle


\begin{abstract}
Panoramic Image Generation (PIG) aims to create coherent images of arbitrary lengths. Most existing methods fall in the joint diffusion paradigm, but their complex and heuristic crop connection designs often limit their ability to achieve multilevel coherence. By deconstructing this challenge into its core components, we find it naturally aligns with next-token prediction, leading us to adopt an autoregressive (AR) paradigm for PIG modeling. However, existing visual AR (VAR) models are limited to fixed-size generation, lacking the capability to produce panoramic images. In this paper, we propose PanoLlama, a novel framework that achieves endless and coherent panorama generation with the autoregressive paradigm. Our approach develops a training-free strategy that utilizes token redirection to overcome the size limitations of existing VAR models, enabling next-crop prediction in both horizontal and vertical directions. This refreshes the PIG pipeline while achieving SOTA performance in coherence (47.50\%), fidelity(28.16\%), and aesthetics (15\%). Additionally, PanoLlama supports applications other PIG methods cannot achieve, including mask-free layout control, multi-scale and multi-guidance synthesis. To facilitate standardized evaluation, we also establish a dataset with 1,000 prompts spanning 100+ themes, providing a new testing benchmark for PIG research. The code is available at \href{https://github.com/0606zt/PanoLlama}{https://github.com/0606zt/PanoLlama}.
\vspace{-1em}
\end{abstract}


\begin{figure*}[!htbp]
    \centering
    \includegraphics[width=\textwidth]{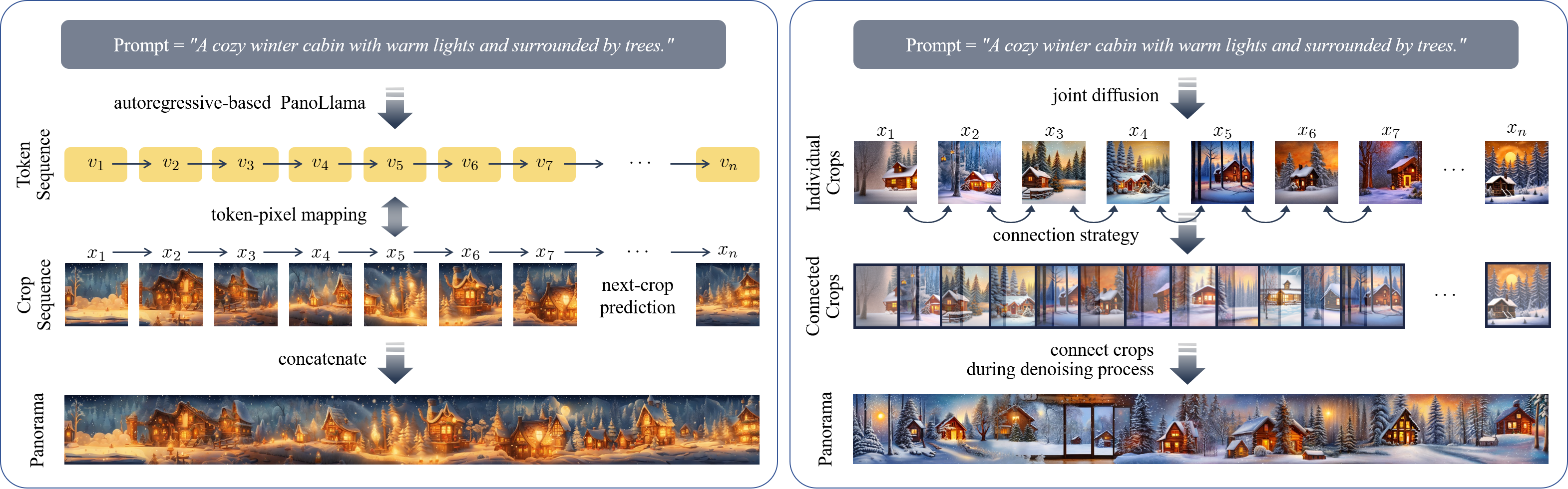}
    \caption{PanoLlama vs. the most effective current PIG method, joint diffusion. Joint diffusion methods (right) denoise each image crop individually and require complex designs to connect them, which face challenges in achieving multilevel coherence. To address this, our PanoLlama (left) redefines PIG as a next-crop prediction task. Leveraging the autoregressive properties that inherently align with the PIG objective, we generate endless panoramas with higher continuity.\vspace{-1em}}
    \label{fig:intro}
\end{figure*}

\section{Introduction}
\label{sec:introduction}

Panoramic Image Generation (PIG) \cite{bar2023multidiffusion} is the task of generating images with arbitrarily long dimensions. It has gained significant attention as a subtask in image generation, especially with the growing interest in artistic expression, user control, and historical restoration \cite{yang2023diffusion}. The objective of this task is to generate coherent and endless panoramas under a planar perspective.

In recent years, diffusion models \cite{sohl2015deep} have become the mainstream method in image generation, proving effective in visual editing \cite{huang2025diffusion}, restoration \cite{li2023diffusion}, augmentation \cite{chen2024comprehensive}, and multi-modal tasks \cite{jiang2024survey,ruan2023mm,hu2025dynamicid}. They employ diffusion and denoising Markov chains to map the data distribution to a Gaussian distribution, as exemplified by DDPM \cite{ho2020denoising}, DDIM \cite{song2021denoising}, and LDM \cite{rombach2022high}. So far, the most effective paradigm for PIG is joint
diffusion \cite{bar2023multidiffusion,lee2023syncdiffusion,quattrini2025merging}, where the panoramic latent space is segmented into individual crops, then denoised and fused into a panorama through the connection strategy (\cref{fig:intro} right). While effective, these approaches face limitations in the following aspects.

\begin{enumerate}
    \item \textbf{Multilevel-Coherence Challenge}: Existing methods often rely on heuristic, rule-based connection strategies, such as performing weighted averaging \cite{bar2023multidiffusion}, introducing gradient guidance \cite{lee2023syncdiffusion,zhou2024twindiffusion}, or deepening attention layers \cite{quattrini2025merging}, struggling to achieve multi-coherence across both low-level and high-level features.

    \item \textbf{Complex Designs}: Achieving crop-wise generation is non-trivial. These methods require complex designs to reconcile denoising paths across image crops, lacking unity, stability, and scalability.
\end{enumerate}

Achieving low-level coherence (e.g., color, edges) relies on adjacent crop connections, while achieving high-level coherence (e.g., semantics, layout) requires considering the overall transition across a sequence of crops, which naturally aligns with the next-token prediction paradigm. This insight motivates us to handle PIG tasks through an autoregressive (AR) architecture, which has been widely used in large language models (LLMs) \cite{vaswani2017attention,brown2020language}. Visual AR (VAR) approaches \cite{yu2024language,bai2024sequential,sun2024autoregressive} introduce a tokenizer to quantize image features into discrete tokens, generating an image in the next-token prediction manner.

However, unlike LLMs' ability to produce long text, existing VAR models are limited to fixed-size image generation (typically \(512\times512\)) constrained by the training paradigm of image data.

To address these challenges, we propose PanoLlama (\cref{fig:intro} left), a novel framework that achieves endless and coherent panorama generation with the autoregressive paradigm. Our key contributions are as follows:

\begin{enumerate}
    \item \textbf{New Paradigm}: We define a new paradigm for PIG, modeling it as a \textbf{next-token prediction} task to better solve the multilevel coherence challenge.

    \item \textbf{New Strategy}: Based on token redirection, we develop a training-free \textbf{next-crop prediction} strategy that enables endless PIG with existing VAR models. Compared to current methods with complex designs, PanoLlama offers a more straightforward and efficient framework, achieving SOTA performance in coherence (47.50\%), fidelity \& diversity (28.16\%), and aesthetics (15\%).

    \item \textbf{Additional Applications}: Beyond basic panorama generation, we support applications other PIG methods cannot achieve, including multi-scale generation, mask-free layout control, and multi-guidance synthesis.

    \item \textbf{New Benchmark}: Given the lack of a standardized testing prompt in prior PIG works, which typically rely on 5-20 specific ones, we construct a dataset of 1,000 detailed prompts across 100+ themes. Along with a comprehensive set of baselines and metrics, this establishes a new benchmark for panorama generation.
\end{enumerate}


\section{Related Work}
\label{sec:related_work}

\paragraph{Panorama Generation} While panoramas often refer to 360° spherical projections \cite{chen2022text2light,wang2023360,wu2024panodiffusion,li2024panogen,ni2025wonderfree}, we define them more broadly as arbitrarily long images, aiming for endless and coherent extension in a planar view. Existing studies mainly focus on extending pre-trained diffusion models for panorama upscaling, which can be summarized into two branches. (i) The main branch is joint diffusion, aiming to develop a connection strategy that seamlessly fuses image crops into a panorama. This branch is pioneered by MultiDiffusion \cite{bar2023multidiffusion}, where the panoramic space is split into smaller crops, then denoised and merged together through a linear interpolation function. Building on this, many studies \cite{lee2023syncdiffusion,zhou2024twindiffusion,quattrini2025merging,lee2024streammultidiffusion,zhang2024multi} are devoted to optimizing the connection process, resulting in improved generation outcomes. However, as discussed in \cref{sec:introduction}, these methods often rely on intuitively designed connection strategies, limiting their ability to achieve multilevel coherence. (ii) Another branch utilizes image inpainting techniques to establish regional correlations for extrapolating subsequent image content \cite{avrahami2023blended,rombach2022high}. Yet, they infer the next crop's content solely from the masked region of the previous one, lacking consideration for the overall layout and structure, and requiring long generation times.

\paragraph{Autoregressive Models for Image Generation} Visual AR models \cite{xiong2024autoregressive} introduce a tokenizer to enable sequential modeling for image data, with its generation process following the next-token prediction paradigm. VQVAE \cite{van2017neural} pioneers this approach by incorporating a codebook to convert continuous image embeddings into discrete tokens. VQGAN \cite{esser2021taming,cao2023efficient} further improves VQVAE by integrating adversarial and perceptual losses. Recent studies \cite{yu2024language,wang2024emu3,chang2022maskgit,tian2024visual,li2024controlvar} attempt to incorporate image generation with LLMs (e.g., GPT \cite{brown2020language}, Gemini \cite{team2023gemini}, and Llama \cite{touvron2023llama}), which have demonstrated their effectiveness in understanding and generating visual content. For example, LlamaGen \cite{sun2024autoregressive} explores the vanilla Llama architecture \cite{touvron2023llama} for image generation and yields promising results. While we propose that the autoregressive paradigm can better tackle the multilevel-coherence problem, current VAR methods lack the capability for panorama generation. This provides insights for our work, which bridges this gap and addresses the key challenge of PIG tasks.


\section{Methodology}

\begin{figure*}[!htbp]
    \centering
    \includegraphics[width=\textwidth]{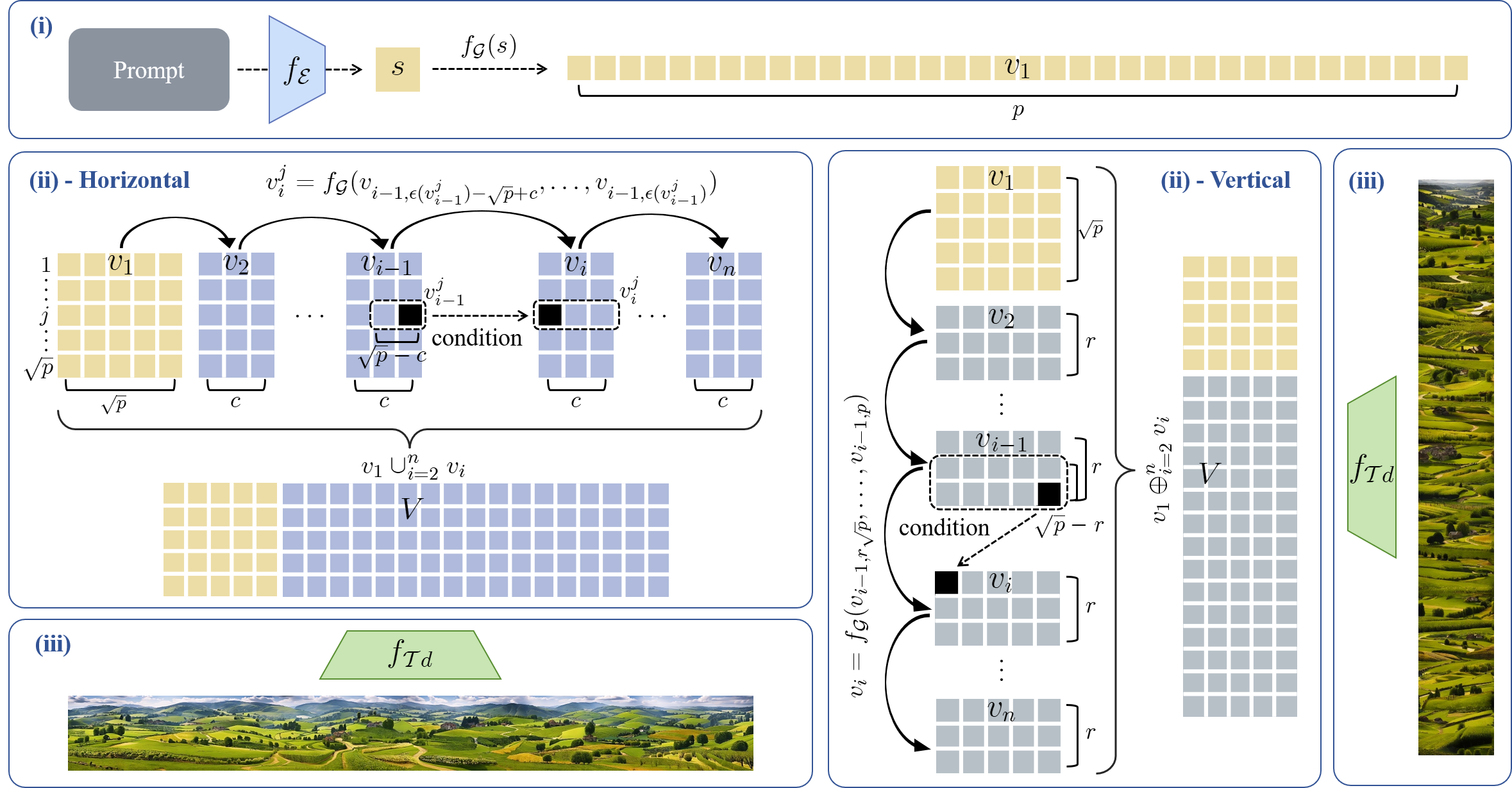}
    \caption{The framework of PanoLlama consists of three parts: (i) \textbf{Textual Conditioning}: The given prompt \(y\) is encoded into a conditional embedding \(s\) using the text encoder \(f_\mathcal{E}\) (\cref{eq:prompt_encode}). (ii) \textbf{Next-Crop Prediction}: With \(s\) as a prefilling, the token generator \(f_\mathcal{G}\) autoregressively generates a sequence of \(p\) image tokens (\cref{eq:fir_seq}). After reaching the token limit, \textbf{vertical expansion} is achieved by setting \(v_i\) to start with the last \(p-r\sqrt{p}\) tokens of \(v_{i-1}\) via \cref{eq:sec_seq_verti}; while \textbf{horizontal expansion} extends rows interleavingly by \cref{eq:sec_seq_horiz}, redirecting the last \(\sqrt{p}-c\) tokens of each preceding row \(v_{i-1}^j\) as the starting condition for \(v_i^j\). (iii) \textbf{Decoding Tokens into Panorama}: After \(n\) rounds of expansion, the concatenated token sequence \(V\) is converted into the panorama \(x'\) through the image tokenizer decoder \(f_{\mathcal{T}d}\) (\cref{eq:tokenizer_decode}).\vspace{-1em}}
    \label{fig:method}
\end{figure*}

\subsection{Panorama Generation as a Next-Token Prediction Task}
\label{sec:formulation}

In PIG tasks, we aim to extend a pre-trained reference model \(\Phi\), which operates within a limited image space \(\mathcal{I}=\mathbb{R}^{h\times w\times ch}\), to enable panorama generation over an arbitrarily large space \(\mathcal{I}'=\mathbb{R}^{h'\times w'\times ch}\), where \(h'>h\) and \(w'>w\). Specifically, the panorama \(x'\in\mathcal{I}'\) is constructed by a series of spatial-ordered image crops \(\mathcal{X}=\{x_i\mid x_i\in\mathcal{I},i=1,2,\ldots,n\}\), where each \(x_i\) represents an individual image segment.

PanoLlama defines a joint probability distribution to learn how these crops interrelate to form a panorama \(x'\):
\begin{equation}\label{eq:joint}
\begin{aligned}
    P(x')
    &=P(x_1,x_2,\ldots,x_n) \\
    &=\prod_{i=1}^{n}P(x_i\mid x_1,x_2,\ldots,x_{i-1})
\end{aligned}
\end{equation}
where each \(P(x_i\mid x_1,x_2,\ldots,x_{i-1})\) represents the conditional probability of generating the \(i\)-th image crop given the preceding ones. This draws from the inspiration of \textbf{next-token prediction} in language modeling, creatively applied to generate a coherent crop series for panorama.

Taking the logarithm of both sides of \cref{eq:joint}, we obtain our loss function:
\begin{equation}\label{eq:loss}
\begin{aligned}
    \mathcal{L}(\theta)
    &=-\log P(x'\mid\theta) \\
    &=-\sum_{i=2}^{n}\log P(x_i\mid x_1,x_2,\ldots,x_{i-1};\theta)
\end{aligned}
\end{equation}
Formally, the optimization task is:
\begin{equation}\label{eq:opti_task}
\begin{aligned}
    \theta^*
    &=\arg\min_{\theta}\mathcal{L}(\theta) \\
    &=\arg\min_{\theta}-\sum_{i=2}^{n}\log P(x_i\mid x_1,x_2,\ldots,x_{i-1};\theta)
\end{aligned}
\end{equation}
By leveraging a pre-trained autoregressive model as \(\Phi\), we can well estimate the \(\theta^*\) that maximizes the likelihood of \(x'\), leading to enhanced multi-coherence of the panorama data.

\subsection{Framework of PanoLlama}
\label{sec:framework}







However, as discussed in \cref{sec:introduction} and~\ref{sec:related_work}, existing VAR models are limited to generating images \(x\in\mathcal{I}\) with the token constraint of \(p\). To address this, we develop a simple yet effective approach that extends next-token prediction within a single crop to \textbf{next-crop prediction} across crops, enabling image generation in a panoramic space \(\mathcal{I'}\).

Given the model's property of arranging image tokens in a raster scan order, we treat each token sequence as a 2D block to achieve next-crop expansion in a row-wise, training-free manner. Distinct strategies are designed for both horizontal and vertical expansion, with the entire process recapped in \cref{fig:method} and \cref{appendix:gen_process}.

Firstly, the textual prompt \(y\) is encoded into an embedding \(s\) using the text encoder function \(f_\mathcal{E}\):
\begin{equation}\label{eq:prompt_encode}
    s=f_\mathcal{E}(y)
\end{equation}
where \(f_\mathcal{E}:\mathcal{Y}\rightarrow\mathcal{E}\) maps the prompt space \(\mathcal{Y}\) to the textual embedding space \(\mathcal{E}\).

With \(s\) as a conditional prefilling, the token generator \(f_\mathcal{G}\) autoregressively produces a sequence of \(p\) image tokens, labeled as the first block:
\begin{equation}\label{eq:fir_seq}
   v_1=f_\mathcal{G}(s)=\{v_{1,k}\mid k=1,2,\ldots,p\}
\end{equation}

\paragraph{Vertical Expansion} When the position index \(k\) reaches the token limit \(p\) for the first block, it is redirected to \(p-r\sqrt{p}\) to start a new round, extending \(r\) rows as \(v_2\). To ensure continuity and maximize token usage, we let \(v_2\) begin with the condition of \(v_{1,r\sqrt{p}},\ldots,v_{1,p}\), i.e., the last \(p-r\sqrt{p}\) tokens of \(v_1\). The generation process then advances as follows:
\begin{equation}\label{eq:sec_seq_verti}
\resizebox{0.9\hsize}{!}{$
   v_2=f_\mathcal{G}(v_{1,r\sqrt{p}},\ldots,v_{1,p})=\{v_{2,k}\mid k=1,2,\ldots,r\sqrt{p}\}
$}
\end{equation}

This redirection process is iteratively applied for \(n\) times, resulting in a series of token sequences \(\mathcal{V}=\{v_i\mid i=2,\ldots,n\}\), with each \(v_i\) corresponds to the image crop \(x_i\). Then the entire panorama is constructed by:
\begin{equation}\label{eq:pan_seq_verti}
\begin{aligned}
    V
    &=v_1\oplus_{i=2}^{n}v_i \\
    &=\{v_{1,k}\mid k=1,2,\ldots,p\} \\
    &\oplus\{v_{i,k}\mid i=2,\ldots,n,\;k=1,2,\ldots,r\sqrt{p}\}
\end{aligned}
\end{equation}
where \(\oplus\) refers to the vertical concatenation operator.

\paragraph{Horizontal Expansion} Unlike vertical expansion, which naturally aligns with the original token arrangement, horizontal expansion disrupts this order. Thus, we propose an interleaved expansion method as an effective solution.

Let \(v_i^j\) denote the \(j\)-th row in the \(i\)-th block, \(\epsilon(v_i^j)\) denote the index of the end token in \(v_i^j\). In the horizontal case, each adjacent row \(v_{i-1}^j\) and \(v_i^j\) need to be connected with successive and cohesive conditions. Thus, we modify the redirection strategy \cref{eq:sec_seq_verti} to let the last \(\sqrt{p}-c\) tokens of each \(v_{i-1}^j\) serves as the start condition for \(v_i^j\). In each iteration, the length of each row is extended by \(c\) columns:
\begin{equation}\label{eq:sec_seq_horiz}
\resizebox{0.88\hsize}{!}{$
\begin{aligned}
   v_i^j
   &=f_\mathcal{G}(v_{i-1,\epsilon(v_{i-1}^j)-\sqrt{p}+c},\ldots,v_{i-1,\epsilon(v_{i-1}^j)}) \\
   &=\{v_{i,k}\mid j=1,2,\ldots,\sqrt{p},\;k=c(j-1)+1,\ldots,cj\}
\end{aligned}
$}
\end{equation}

Accordingly, \cref{eq:pan_seq_verti} should be modified as:
\begin{equation}\label{eq:pan_seq_horiz}
   \begin{aligned}
    V
    &=v_1\cup_{i=2}^{n}v_i \\
    &=\{v_{1,k}\mid k=1,2,\ldots,p\} \\
    &\cup\{v_{i,k}\mid i=2,\ldots,n,\;k=1,2,\ldots,c\sqrt{p}\}
\end{aligned}
\end{equation}
and \(\cup\) corresponds to the horizontal concatenation operator.

After vertical / horizontal expansion, we convert \(V\) into a panorama \(x'\) through the decoder of image tokenizer \(f_{\mathcal{T}d}\):
\begin{equation}\label{eq:tokenizer_decode}
   x'=f_{\mathcal{T}d}(V)
\end{equation}
where \(f_{\mathcal{T}d}:\mathcal{T}\rightarrow\mathcal{I'} \) maps tokens to the pixel space, and \(\mathcal{T}\) denotes the discrete token set.

\subsection{PanoLlama vs. Existing PIG Methods}
\label{sec:ours_vs_others}

As outlined in \cref{sec:related_work}, current PIG methods fall into two branches: joint diffusion and inpainting. To provide a more in-depth comparison and analysis of PanoLlama against these methods, we unify their paradigms into the modeling form presented by \cref{sec:formulation}.

\paragraph{PanoLlama vs. Inpainting} For the inpainting strategy, each crop \(x_i\) acts as the masked part of the previous one, inferring content from it. The conditional dependency can be expressed as:
\begin{equation}\label{eq:joint_inpainting}
    P(x')=\prod_{i=2}^{n} P(x_i\mid x_{i-1})
\end{equation}

This can be viewed as a simplification of our paradigm in \cref{eq:joint}, considering features only from the left neighbor \(x_{i-1}\), whereas our approach accounts for all preceding crops \(x_1,x_2,\ldots,x_{i-1}\), facilitating better coherence.

\paragraph{PanoLlama vs. Joint Diffusion} The key to joint diffusion is devising an effective connection strategy between adjacent crops. This leads to the joint probability:
\begin{equation}\label{eq:joint_sd}
    P(x')=\prod_{i=2}^{n} P(x_i\mid x_{i-1},x_{i+1})
\end{equation}
where each crop \(x_i\) is generated based on its left and right neighbor.

While existing joint diffusion methods have explored a range of connection strategies for this task, their heuristic and tailored designs face difficulties in achieving multi-coherence holistically. In contrast, our approach leverages the autoregressive properties that inherently align with the PIG objective, forming an effective framework that surpasses these methods.


\begin{figure*}[!htbp]
    \centering
    \includegraphics[width=\textwidth]{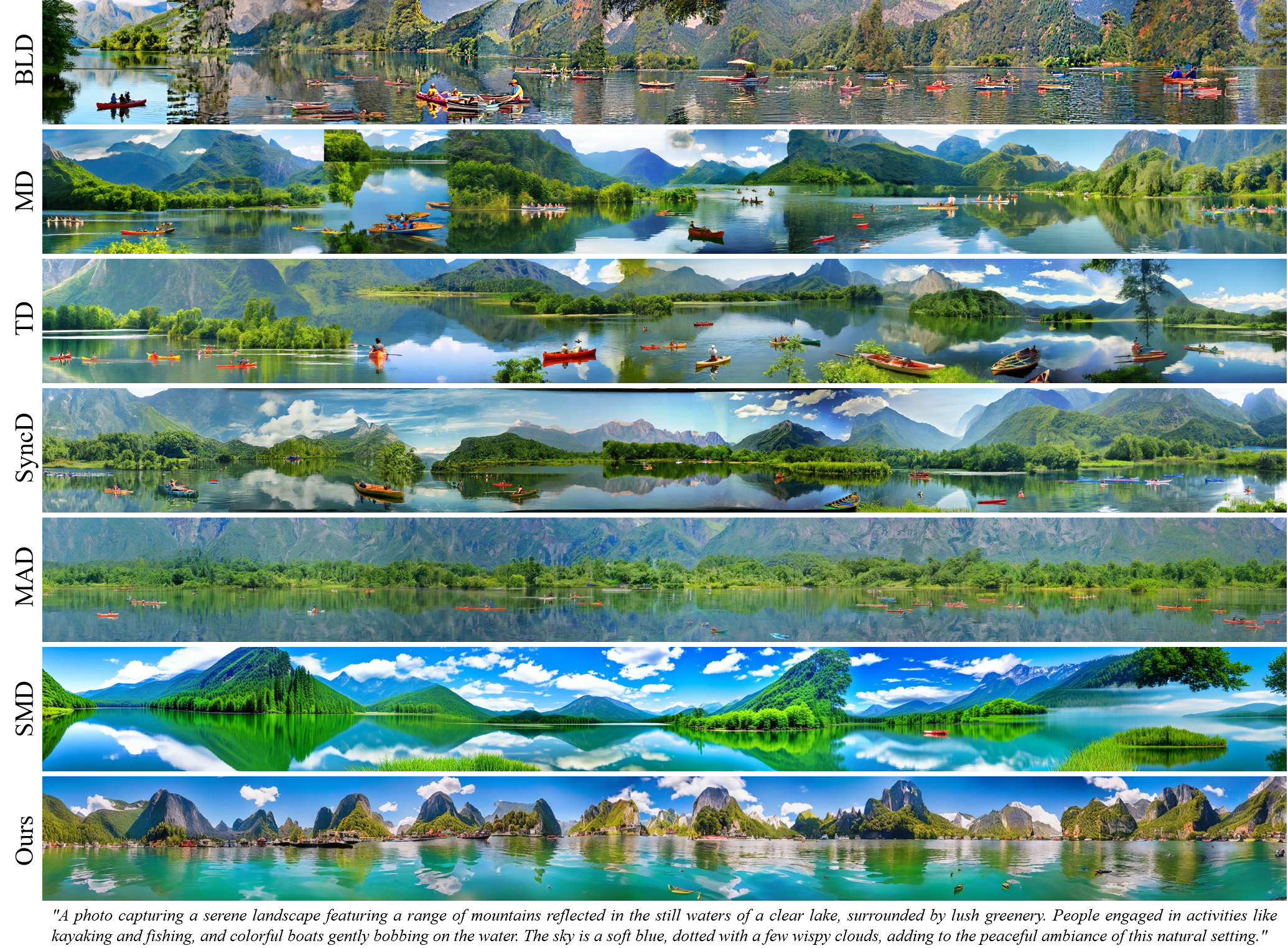}
    \caption{Qualitative comparisons between six baselines and ours. Our approach minimizes unnatural connections commonly found in existing methods, producing high-quality panoramas with significantly extended dimensions of \(512\times5120\).\vspace{-1em}}
    \label{fig:qual_comp}
\end{figure*}

\section{Experiments}

We conduct a comprehensive evaluation of PanoLlama, comparing it with multiple baselines from both qualitative and quantitative perspectives. Since some baselines do not support vertical generation, the experiments discussed in \cref{sec:comparison} and \ref{sec:ablation} focus on the horizontal case.

\subsection{Comparison}
\label{sec:comparison}

\paragraph{Evaluation Dataset} To ensure a comprehensive and fair evaluation, we design the dataset with a focus on diversity and difficulty across content types. It contains 1,000 detailed long prompts for text-to-panorama generation, covering 25 main themes and 100+ sub-themes, encompassing both photorealistic and artistic styles. See \cref{appendix:dataset} for more construction details.

\paragraph{Implementation Details} Using all prompts in the dataset and 25 sets of random seeds, we generate 2,000 panoramas in total. Specifically, these panoramas are produced with an expansion stride of \(u=\frac{3}{4}\) and dimensions of \(h'=512, w'=5120\), 10x wider than the original \(\Phi\).

\begin{table*}[!htbp]
    \fontsize{12pt}{17pt}\selectfont
    \centering
    \caption{Quantitative comparison of PanoLlama with multiple baselines. The best, second-best, and relative scores are marked in bold, underlined, and parentheses. Our method achieves the best overall performance, excelling in coherence, fidelity, and aesthetics.}
    \label{tab:quant_comp}
    \begin{adjustbox}{width=\textwidth}
	\begin{tabular}{cccccccccc}
            \hline
            \multirow{2}{*}{} & \multicolumn{4}{c}{Coherence} & \multicolumn{2}{c}{Fidelity \& Diversity} & \multicolumn{2}{c}{Compatibility} & Efficiency \\
            \cmidrule(lr){2-5} \cmidrule(lr){6-7} \cmidrule(lr){8-9} \cmidrule(lr){10-10}  
    	& LPIPS\(\downarrow\) & DISTS\(\downarrow\) & TV\(\downarrow\) & SSIM\(\uparrow\) & FID\(\downarrow\) & IS\(\uparrow\) & CLIP\(\uparrow\) & CLIP-aesthetic\(\uparrow\) & time\(\downarrow\) \\
    	\hline
            SD & -- & -- & -- & -- & 36.14 & 7.49 & 33.03 & 6.74 & -- \\
            LlamaGen & -- & -- & -- & -- & 37.82 & 6.43 & 31.62 & 6.74 & -- \\
            \hdashline
            BLD & 0.829 & 0.369 & 0.083 & 0.008 & 89.51 (+53.37) & 5.81 (-1.68) & 32.29 (-0.74) & 5.76 (-0.98) & 10670.35s \\
            MD & 0.694 & 0.269 & 0.061 & 0.184 & \underline{39.30 (+3.16)} & 6.36 (-1.13) & 33.19 (+0.16) & 6.84 (+0.10) & 1808.95s \\
            TD & 0.651 & 0.239 & 0.059 & 0.261 & 42.10 (+5.96) & 6.22 (-1.27) & \textbf{33.78 (+0.75)} & 6.88 (+0.14) & 2045.79s \\
    	SyncD & 0.582 & 0.231 & 0.058 & 0.263 & 44.89 (+8.75) & 6.13 (-1.36) & \underline{33.21 (+0.18)} & \underline{6.94 (+0.20)} & 7232.70s \\
    	MAD & \underline{0.520} & \underline{0.209} & \underline{0.040} & \underline{0.268} & 59.23 (+23.09) & 4.99 (-2.50) & 31.30 (-1.73) & 6.90 (+0.16) & 1923.99s \\
            SMD & 0.637 & 0.223 & 0.055 & 0.257 & 89.64 (+53.50) & \underline{6.45 (-1.04)} & 30.95 (-2.08) & 6.75 (+0.01) & \textbf{241.06s} \\
            Ours & \textbf{0.410} & \textbf{0.196} & \textbf{0.021} & \textbf{0.305} & \textbf{40.09 (+2.27)} & \textbf{5.97 (-0.46)} & 31.56 (-0.06) & \textbf{6.97 (+0.23)} & \underline{726.33s}\vspace{0.04cm} \\
    	\hline
	\end{tabular}
 \end{adjustbox}
\end{table*}\vspace{-1em}

\paragraph{Evaluation Metrics} We apply nine quantitative metrics to evaluate our approach across four aspects: (i) coherence at crop connections, (ii) fidelity and diversity of generated panoramas, (iii) compatibility with input prompts, and (iv) efficiency of the inference process. Since the performance of PIG methods is closely tied to their reference model \(\Phi\), we also measure \(\Phi\)'s performance on (ii) and (iii) as a reference score (\,(i) and (iv) are panorama-specific and cannot be evaluated for \(\Phi\)), a common practice in related studies \cite{bar2023multidiffusion,lee2023syncdiffusion,zhou2024twindiffusion}. The score gap between the PIG method and its corresponding \(\Phi\) reflects the degradation (or improvement) introduced during panorama expansion.

\begin{itemize}
    \item \textbf{\underline{Coherence}}: We use LPIPS \cite{zhang2018unreasonable} and DISTS \cite{ding2020image} to capture perceptual differences of high-level features; SSIM \cite{wang2003multiscale} and TV \cite{rudin1992nonlinear} to measure pixel differences for low-level features. The first three metrics are calculated between pairs of adjacent but non-overlapping crops, and the TV score is computed along the connection between crops.

    \item \textbf{\underline{Fidelity \& Diversity}}: FID \cite{heusel2017gans} and IS \cite{salimans2016improved} describe both fidelity and variety of the results. We randomly take one crop of each panorama in a size aligned with \(\Phi\), forming a generated image set \(\mathcal{D}_x\). We correspondingly construct two reference sets \(\mathcal{D}_\Phi^1\) and \(\mathcal{D}_\Phi^2\), each with an equal number of images generated by the reference model \(\Phi\). For PIG methods, IS is computed on \(\mathcal{D}_x\), and FID compares distributions of \(\mathcal{D}_x\) with \(\mathcal{D}_\Phi^2\). For \(\Phi\) models, IS is measured on \(\mathcal{D}_\Phi^1\), and FID compares \(\mathcal{D}_\Phi^1\) with \(\mathcal{D}_\Phi^2\).

    \item \textbf{\underline{Compatibility}}: CLIP \cite{radford2021learning} assesses the compatibility between input prompts and generated images, while CLIP-aesthetic \cite{schuhmann2022laion} reflects the overall aesthetic quality. We also use the aforementioned image sets \(\mathcal{D}_x\) and \(\mathcal{D}_\Phi^1\) to calculate these scores.

    \item \textbf{\underline{Efficiency}}: We measure the model's runtime speed by the inference time taken to generate a batch of 50 panoramas on an A100 GPU.
\end{itemize}

\paragraph{Baselines and Reference Models} We compare our PanoLlama with six recent methods: Blended Latent Diffusion (BLD) \cite{avrahami2023blended}, MultiDiffusion (MD) \cite{bar2023multidiffusion}, TwinDiffusion (TD) \cite{zhou2024twindiffusion}, SyncDiffusion (SyncD) \cite{lee2023syncdiffusion}, Merge-Attend-Diffuse (MAD) \cite{quattrini2025merging}, and StreamMultiDiffusion (SMD) \cite{lee2024streammultidiffusion}, using default optimal parameters for each, where \(u\) is typically \(\frac{1}{4}\). As for the reference model \(\Phi\), we take the widely used VAR architecture LlamaGen \cite{sun2024autoregressive}, while other baselines utilize Stable Diffusion (SD) \cite{rombach2022high}. More comparisons with its variants, such as Stable Diffusion XL (SDXL) \cite{podell2024sdxl}, are provided in \cref{appendix:phi_abl}.

\paragraph{Qualitative Results} As shown in \cref{fig:qual_comp}, BLD exhibits artificial seams and repetitive segments during its extrapolation process. As the basic joint diffusion method, MD also suffers from abrupt transitions between crops. While TD and SyncD improve on this, they still show unnatural junctions and layouts. MAD achieves seamless results, but with excessive content repetition and blurred textural details. SMD obtains smoother transitions at the cost of simplifying image content, which results in a loss of features and unstable performance. In contrast, our method consistently generates high-quality panoramas even at 10x extended dimensions of the original \(\Phi\), preserving diversity and detail while demonstrating significantly higher coherence. See \cref{appendix:qual_comp} for additional qualitative results.

\paragraph{Quantitative Results} \cref{tab:quant_comp} reports the mean scores for each metric, with the best in bold and the second-best underlined. The first two rows present results for the reference models, followed by the baselines and our method. As previously mentioned, aspects (i) and (iv) consider absolute scores, while (ii) and (iii) compare relative scores against the reference model, as labeled in parentheses. PanoLlama outperforms baselines in almost all aspects. Coherence, the key factor in panoramic images, is greatly enhanced by our method. Compared to the best baseline MAD, we achieve improvements of 21.15\% in LPIPS, 6.22\% in DISTS, 47.50\% in TV, and 13.81\% in SSIM. An improvement in the CLIP-aesthetic metric is also observed, indicating that the enhanced coherence contributes to better aesthetic quality. As for fidelity and diversity, PanoLlama also achieves the best results, demonstrating that our next-crop prediction strategy introduces less relative loss during the panoramic expansion process. Regarding efficiency, except for the real-time generation method SMD, our generation speed is 3x- 10x faster than other PIG frameworks.

\subsection{Ablation}
\label{sec:ablation}

Our ablation studies further investigate the key factors of PanoLlama's performance, demonstrating its advantages in robustness, adaptability, etc.

As coherence is the most critical attribute for panoramas, we select LPIPS, DISTS, TV, and SSIM metrics for our ablation analysis, construsting a coherence score (COH) based on their weighted sum. Specifically, we replace SSIM with \(1-\)SSIM to ensure monotonicity and normalize all scores to the same range before weighting.\vspace{-0.1em}

\paragraph{Expansion Stride} As described in \cref{sec:framework}, each horizontal iteration expands the 2D token block by \(c\) columns, creating an overlapping region of \(\sqrt{p}-c\) columns between \(x_{i-1}\) and \(x_i\). This yields an expansion stride of \(u=\frac{c}{\sqrt{p}}\), which controls the balance between quality and efficiency. Higher coherence comes with a smaller \(u\), but at the cost of longer generation time. To investigate the effectiveness of our approach in improving this trade-off, we record the COH score and generation time under \(u=1,\frac{3}{4},\frac{1}{2},\frac{1}{4},\frac{1}{8}\), and plot the quality-efficiency curves for different methods.

\begin{figure}[!htbp]
    \centering
    \begin{subfigure}[b]{\linewidth}
        \centering
        \includegraphics[width=\linewidth]{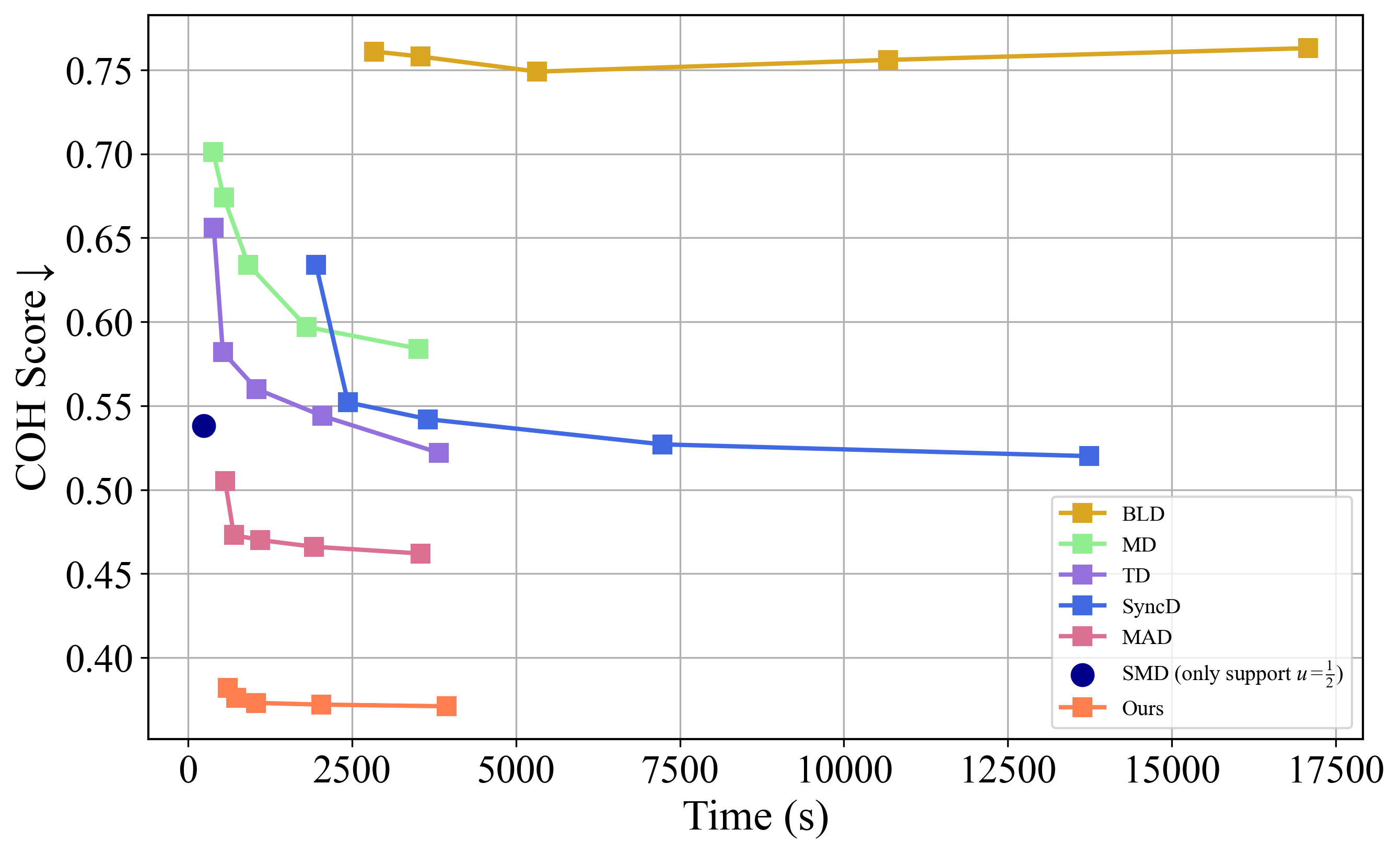}
        \caption{Quality-efficiency curves for different methods, with markers corresponding to \(u=1,\frac{3}{4},\frac{1}{2},\frac{1}{4},\frac{1}{8}\) from left to right. PanoLlama achieves the optimal quality-efficiency trade-off, with less generation time and consistently low COH score.}
        \label{fig:stride_quan_abl}
    \end{subfigure}
    \begin{subfigure}[b]{\linewidth}
        \centering
        \includegraphics[width=\linewidth]{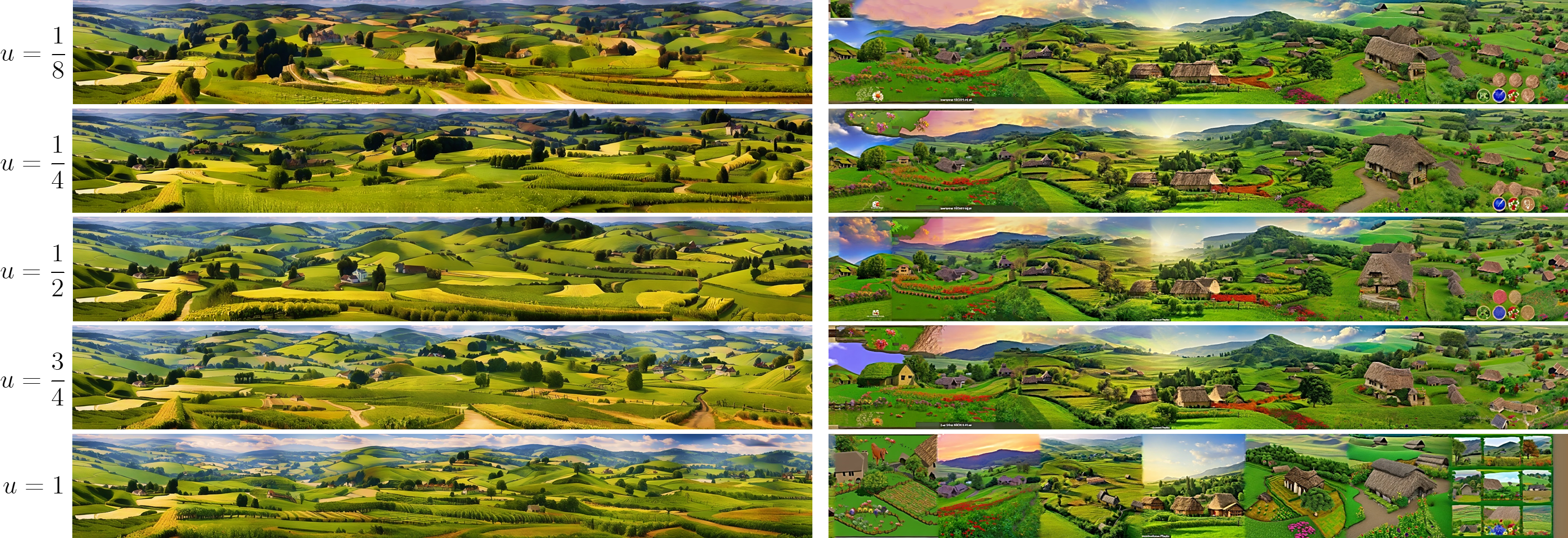}
        \caption{Visual support for the quantitative results. As \(u\) increases, our method (left) shows minimal quality degradation, while seams in the baseline (right) become increasingly noticeable.}
        \label{fig:stride_qual_abl}
    \end{subfigure}
    \caption{The quantitative and qualitative results demonstrate our method's high robustness to the expansion stride \(u\), leading to a better quality-efficiency trade-off.\vspace{-2em}}
    \label{fig:stride_abl}
\end{figure}

As seen in \cref{fig:stride_quan_abl}, at the same value of \(u\), our method's runtime is comparable to other training-free baselines (MD, TD, MAD). However, as \(u\) increases, the performance of other baselines degrades significantly, while our method maintains a consistently low COH score, demonstrating superior robustness and insensitivity to the expansion degree. This results in a superior quality-efficiency balance, with additional visual support presented in \cref{fig:stride_qual_abl}.\vspace{0.5em}

\begin{figure}[!htbp]
    \centering
    \includegraphics[width=\linewidth]{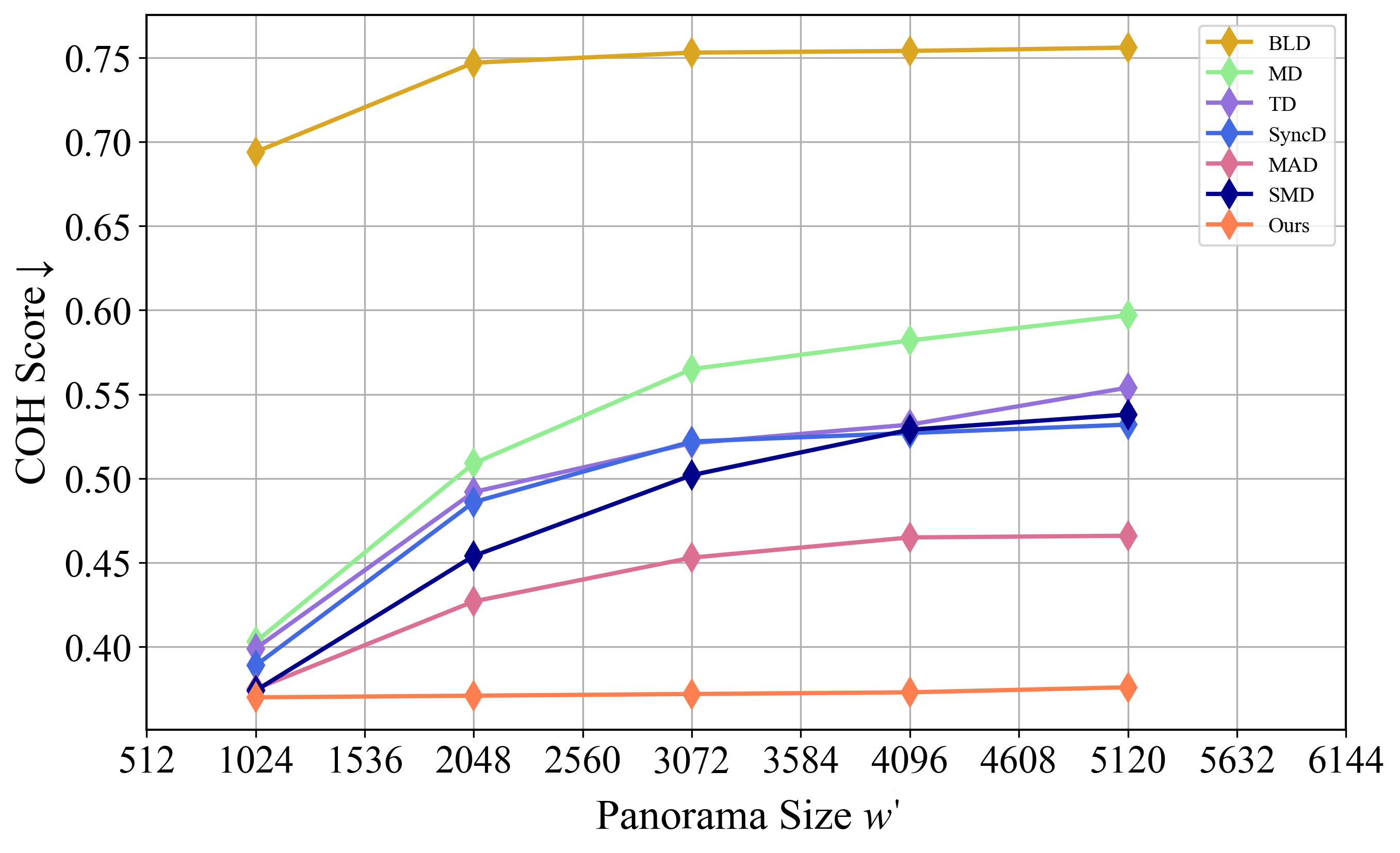}
    \caption{Ablation studies on different panorama sizes \(w'\). While other PIG methods degrade as \(w'\) increases, our PanoLlama maintains stable coherence, effectively handling larger panoramas.\vspace{-2em}}
    \label{fig:size_abl}
\end{figure}

\paragraph{Panorama Size} Generating larger panoramas inherently increases the number of crop connections, making it more challenging to maintain coherence. We test five different sizes of \(w'=1024,2048,3072,4096,5120\), corresponding to 2x, 4x, 6x, 8x, and 10x the resolution of \(\Phi\), to examine PanoLlama's capability in maintaining coherence as panorama length increases.

As shown in \cref{fig:size_abl}, except for BLD, other methods exhibit comparable performance when the size is doubled on \(\Phi\) (i.e., \(w'=1024\)). However, as \(w'\) scales up, other PIG methods become less satisfactory, while our PanoLlama remains stable, effectively reducing the increased potential for incoherence in larger panoramas.

\begin{figure}[!htbp]
    \centering
    \includegraphics[width=\linewidth]{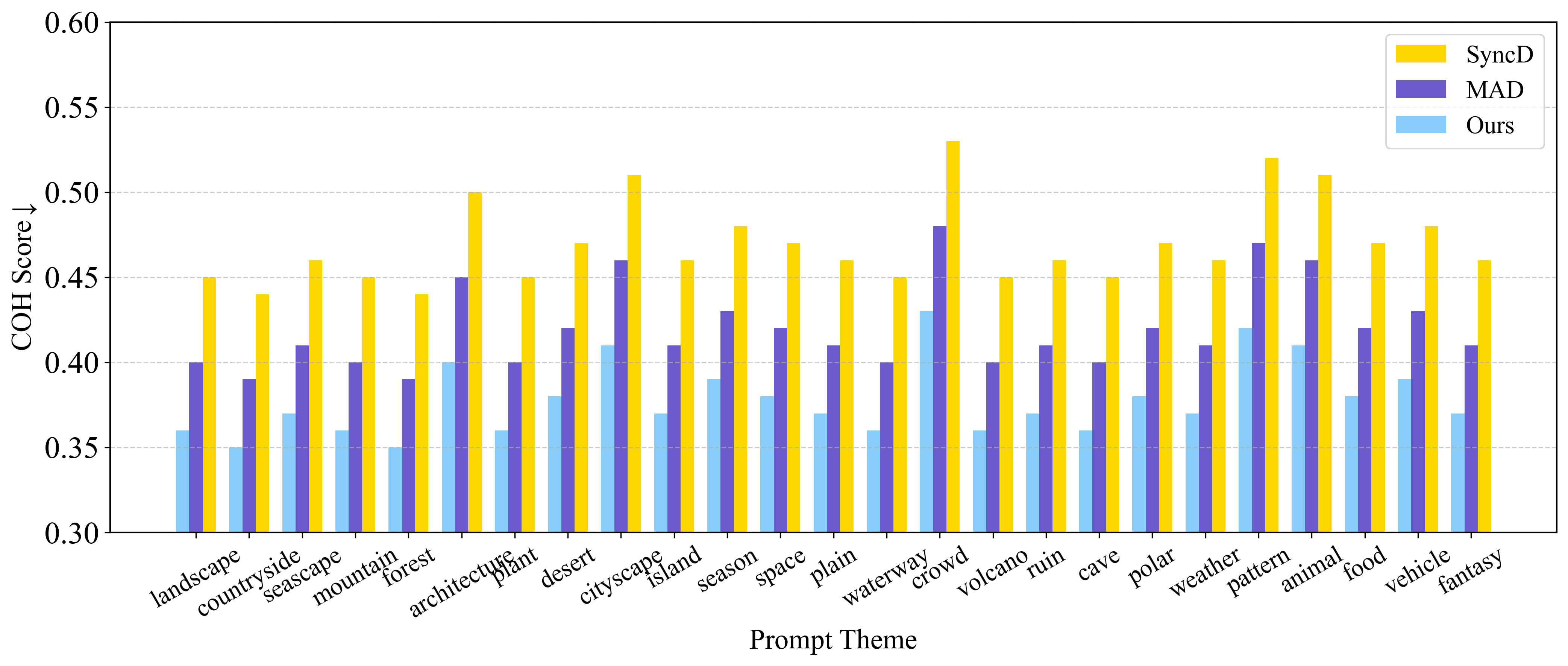}
    \caption{Exploration on a wide range of prompt themes. Our method outperforms MAD and SyncD overall, excelling in grand scenes but facing challenges in complex, dense scenes.\vspace{-2em}}
    \label{fig:theme_abl}
\end{figure}

\paragraph{Prompt Theme} We test each of the 25 main themes in our dataset, generating 80 prompts per theme, to explore which prompt content types pose the greatest challenge for panorama generation.

To streamline, \cref{fig:theme_abl} only displays comparison results with two best-performing baselines, MAD and SyncD. Our method achieves a better overall COH score, exhibiting greater adaptability to prompt contents. Specifically, it often performs well in grand scenes with large, uniform color areas (e.g., 'seascape', 'grassland', 'weather'), but is weaker in complex, dense scenes (e.g., 'crowd', 'pattern').

\begin{figure*}[!htbp]
    \centering
    \includegraphics[width=\textwidth]{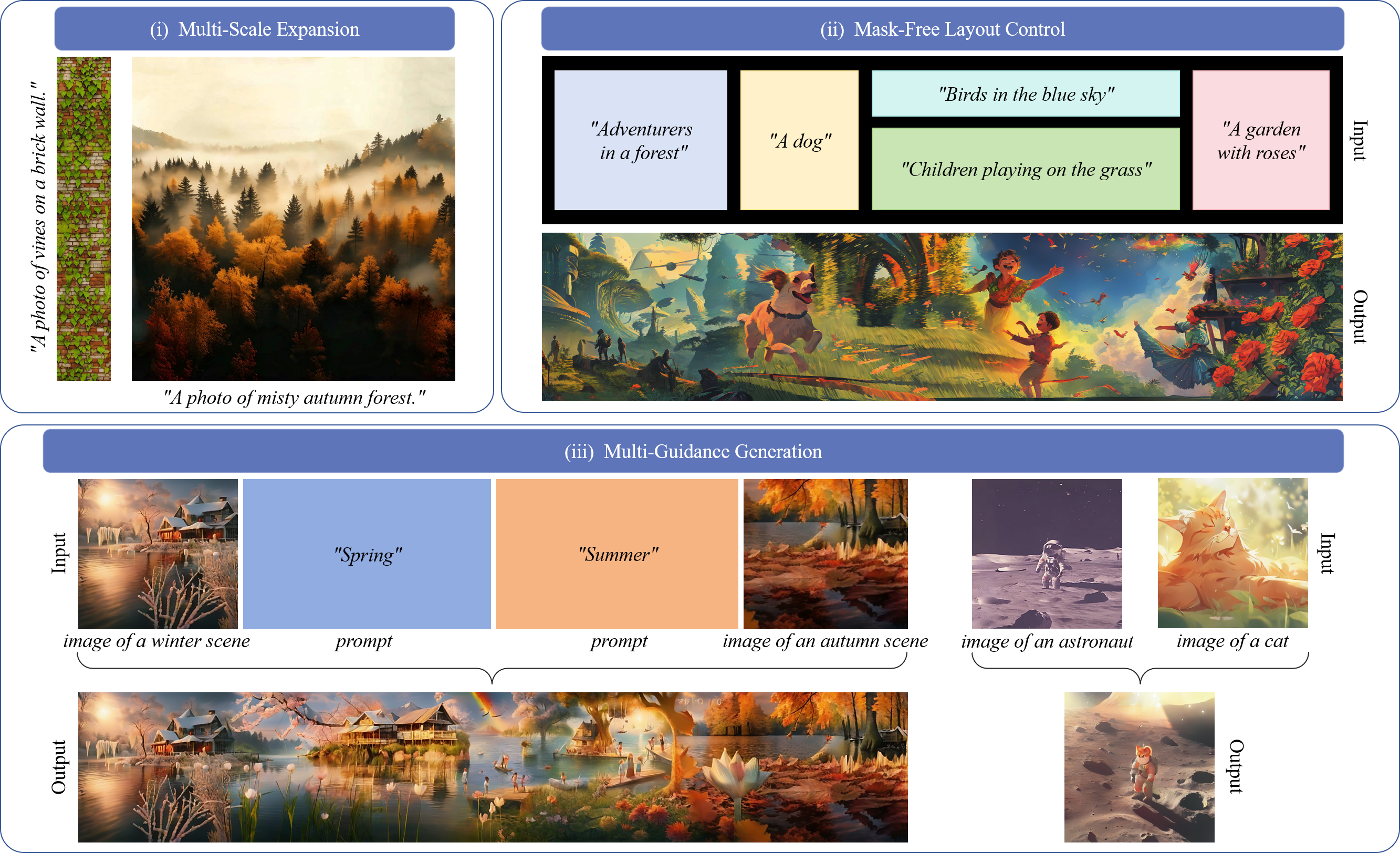}
    \caption{PanoLlama enables versatile applications that other PIG methods cannot achieve, including multi-scale expansion, mask-free layout control, and multi-guidance generation.\vspace{-1em}}
    \label{fig:apps}
\end{figure*}

\subsection{Applications}

Beyond basic panorama generation, we implement a range of applications that other PIG methods cannot achieve, with implementation details provided in \cref{appendix:applications}.

\paragraph{Multi-Scale Expansion} We support expansion in horizontal, vertical, and both directions. \cref{fig:apps} shows the multi-scale results, with sizes of \(3072\times512\) and \(3072\times3072\).

\paragraph{Mask-Free Layout Control} Our method allows for flexible adjustments in rows (\(r\)), columns (\(c\)), and prompts (\(y\)), enabling mask-free layout control. As demonstrated in \cref{fig:apps}, PanoLlama effectively blends different token regions guided by different prompts into a cohesive, storytelling panorama.

\paragraph{Multi-Guidance Generation} Besides text-guided generation, our framework also accommodates image guidance and mixed guidance. Input images are tokenized by the encoder of image tokenizer \(f_{\mathcal{T}e}\), which then serves as the condition for generation. \cref{fig:apps} illustrates two examples where the text and image collaboratively guide the generation process, resulting in an effect similar to image inpainting.


\section{Conclusion} Multilevel coherence is one of the major pending challenges in panoramic image generation. In this paper, we introduce PanoLlama as a novel solution, which has several key advantages over the existing methods: (i) We introduce the next-token prediction paradigm to PIG, offering a better alignment with the core characteristics of multilevel coherence. (ii) We use token redirection to develop a training-free next-crop prediction method, achieving endless PIG upon fixed-size VAR models, with comprehensive experiments demonstrating its SOTA performance in coherence, fidelity, aesthetics, and more. (iii) We support additional applications that other PIG approaches do not achieve. (iv) We establish a new dataset and evaluation benchmark for standardized PIG research.

\paragraph{Limitations} Due to the fixed token capacity of pre-trained VAR models, we approximate global dependencies by leveraging a partial set of preceding tokens, balancing context range with architectural constraints. While this does not fully realize the theoretical ideal of global conditioning, it still achieves strong empirical performance. We leave the extension to full-context conditioning for future work.


\section*{Acknowledgements}

This work is sponsored by Ningbo "Yongjiang Science Innovation 2035" Key Technology Breakthrough Plan (No. 2024Z254). We thank Yunhao Chen for his key contributions to this study, including idea development and coding.


{
    \small
    \bibliographystyle{ieeenat_fullname}
    \bibliography{mybib}
}


\clearpage
\appendix
\onecolumn


\setcounter{page}{1}
\setcounter{figure}{0}
\setcounter{table}{0}
\renewcommand\thefigure{A\arabic{figure}}
\renewcommand\thetable{A\arabic{table}}

\section{PanoLlama Generation Process}
\label{appendix:gen_process}

We provide detailed pseudocode in \cref{alg:gen_process} to facilitate a better understanding of our framework. Aligned with LlamaGen \cite{sun2024autoregressive}, it incorporates FLAN-T5 XL \cite{chung2024scaling} as the text encoder \(f_\mathcal{E}\), VQVAE \cite{van2017neural} as the image tokenizer \(f_\mathcal{T}\), and Llama XL \cite{touvron2023llama} as the token generator \(f_\mathcal{G}\) to achieve text-guided panorama generation.

\begin{algorithm}
\caption{PanoLlama Generation Process}
\label{alg:gen_process}
\begin{algorithmic} 
\State \textbf{Input:} \(\;f_\mathcal{E},f_\mathcal{G},f_{\mathcal{T}d}\quad\!\triangleright\) pre-trained models: text encoder, token generator, decoder of image tokenizer
\State \(\quad\quad\quad\; p\quad\triangleright\) max token limit of image tokenizer
\State \(\quad\quad\quad\; y\quad\triangleright\) textual prompt
\State \(\quad\quad\quad\; mode\quad\triangleright\) direction of expansion
\State \(\quad\quad\quad\; n\quad\triangleright\) expansion iterations
\State \(\quad\quad\quad\; r,c\quad\triangleright\) rows and columns expanded per iteration
\State \textbf{Output:} \(\;x'\quad\triangleright\) panorama
\vspace{0.65em}
\Function{GenerateTokens-Vertical}{$v_{i-1}$, $r$}
    \State \(v_i\gets f_\mathcal{G}(v_{i-1,r\sqrt{p}},\ldots,v_{i-1,p})\)
\State \Return \(v_i\)
\EndFunction
\vspace{0.65em}
\Function{GenerateTokens-Horizontal}{$v_{i-1}$, $c$}
    \For{\(j=1,2,\ldots,\sqrt{p}\)}
        \State \(v_i^j\gets f_\mathcal{G}(v_{i-1,\epsilon(v_{i-1}^j)-\sqrt{p}+c},\ldots,v_{i-1,\epsilon(v_{i-1}^j)})\)
    \EndFor
\State \Return \(v_i\)
\EndFunction
\vspace{0.65em}
\State \(s\gets f_\mathcal{E}(y)\qquad\qquad\qquad\qquad\qquad\qquad\qquad\qquad\qquad\qquad\qquad\;\text{// (i) Textual Conditioning}\)
\vspace{0.65em}
\State \(v_1\gets f_\mathcal{G}(s)\qquad\qquad\qquad\qquad\qquad\qquad\qquad\qquad\qquad\qquad\qquad\text{// (ii) Next-Crop Prediction}\)
\If{\(mode==\text{'Vertical'}\)}
    \For{\(i=2,3,\ldots,n\)}
        \State \(v_i\gets\Call{GenerateTokens-Vertical}{v_{i-1}, r}\)
    \EndFor
    \State \(V\gets v_1\oplus_{i=2}^{n}v_i\)
\ElsIf{\(mode==\text{'Horizontal'}\)}
    \For{\(i=2,3,\ldots,n\)}
        \State \(v_i\gets\Call{GenerateTokens-Horizontal}{v_{i-1}, c}\)
    \EndFor
    \State \(V\gets v_1\cup_{i=2}^{n}v_i\)
\EndIf
\vspace{0.65em}
\State \(x'\gets f_{\mathcal{T}d}(V)\qquad\qquad\qquad\qquad\qquad\qquad\qquad\qquad\qquad\qquad\quad\text{// (iii) Decoding Tokens into Panorama}\)
\State \textbf{Return} \(x'\)
\end{algorithmic}
\end{algorithm}

\clearpage
\section{More Qualitative Comparison Results}
\label{appendix:qual_comp}

\begin{figure}[!htbp]
    \centering
    \includegraphics[width=\linewidth]{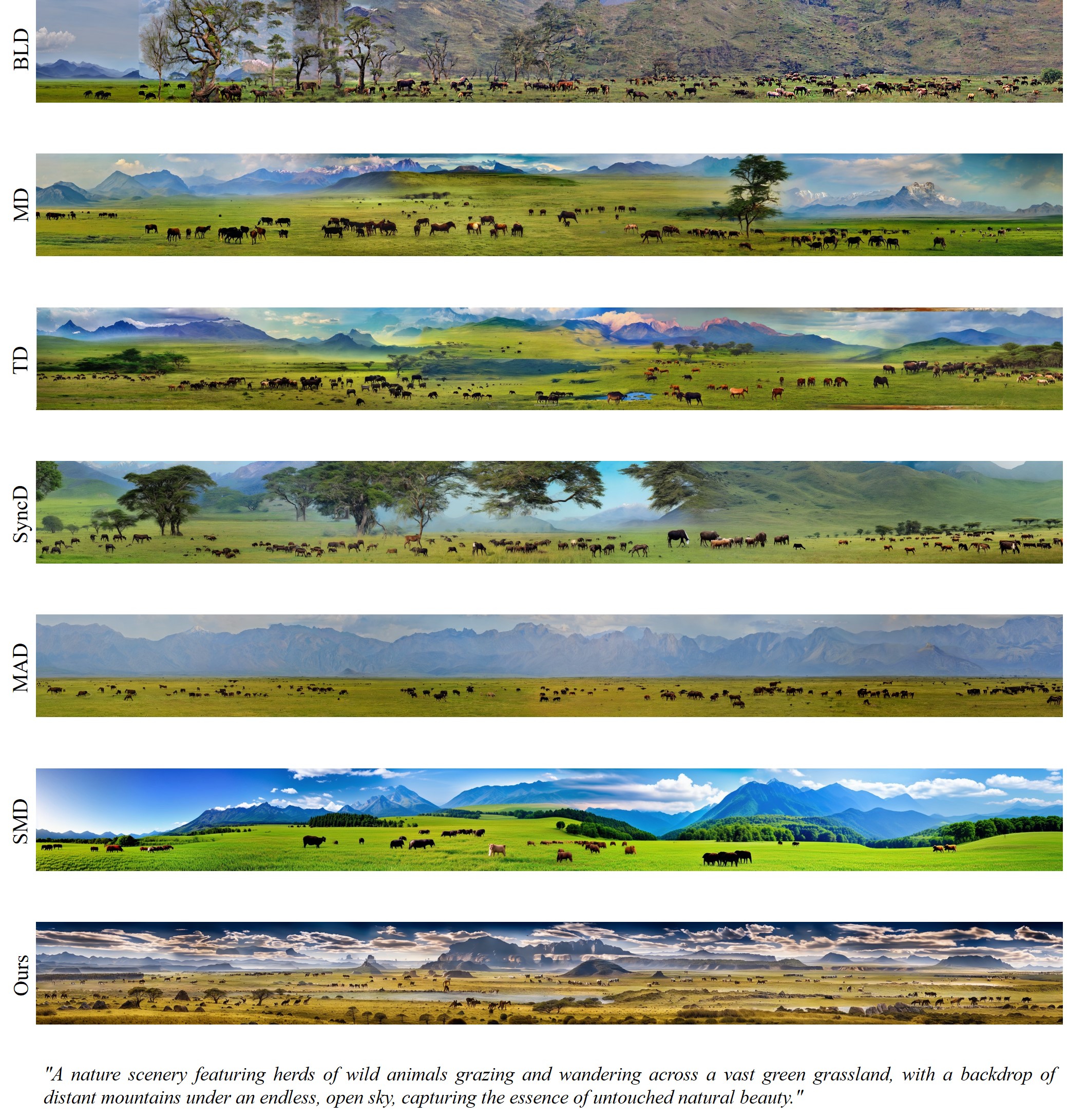}
    \caption{Additional qualitative comparison results on \(512\times5120\) panorama generation.}
    \label{fig:qual_comp_plus_0}
\end{figure}

\begin{figure}[!htbp]
    \centering
    \includegraphics[width=\linewidth]{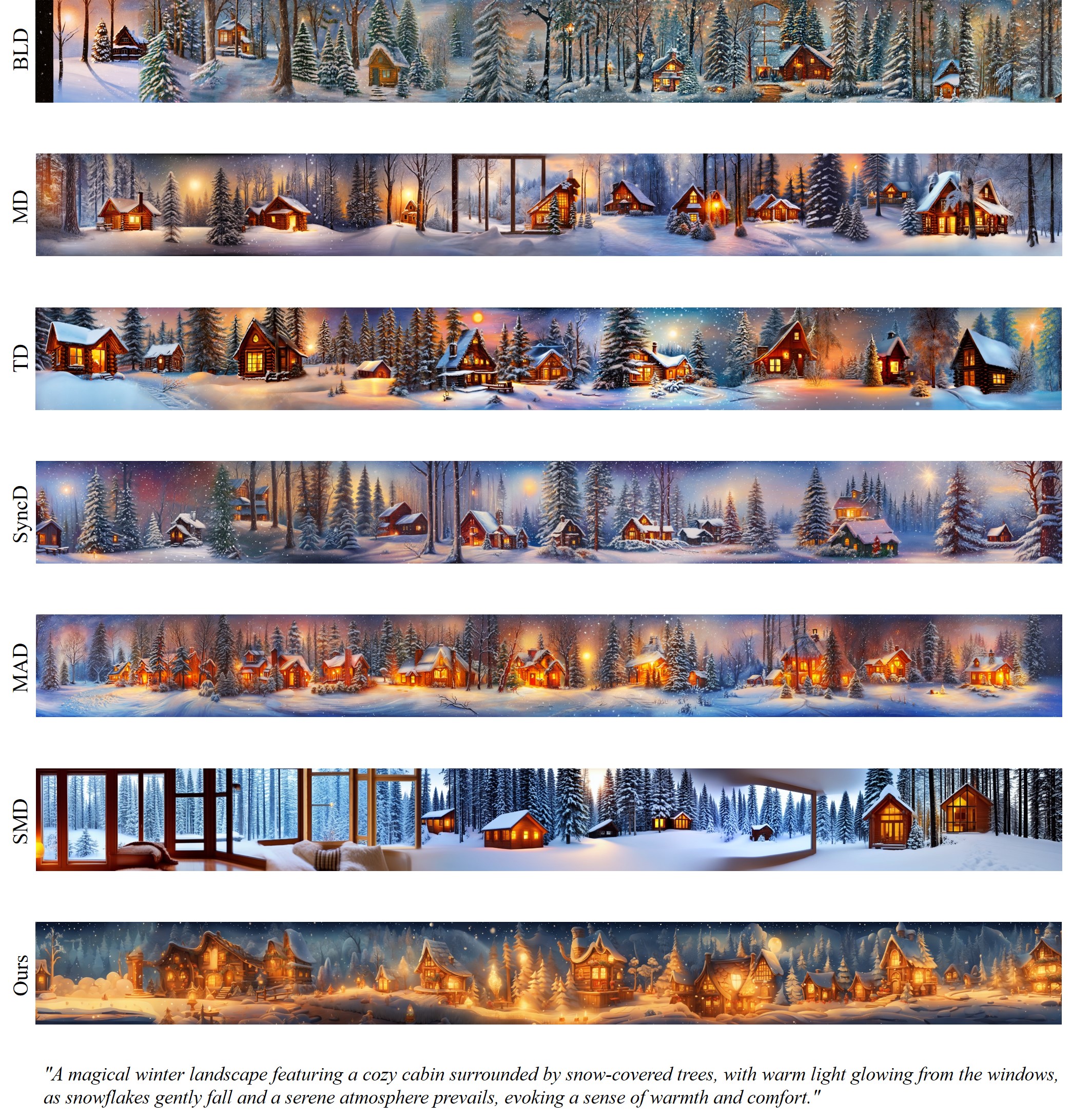}
    \caption{Additional qualitative comparison results on \(512\times5120\) panorama generation.}
    \label{fig:qual_comp_plus_1}
\end{figure}

\begin{figure}[!htbp]
    \centering
    \includegraphics[width=\linewidth]{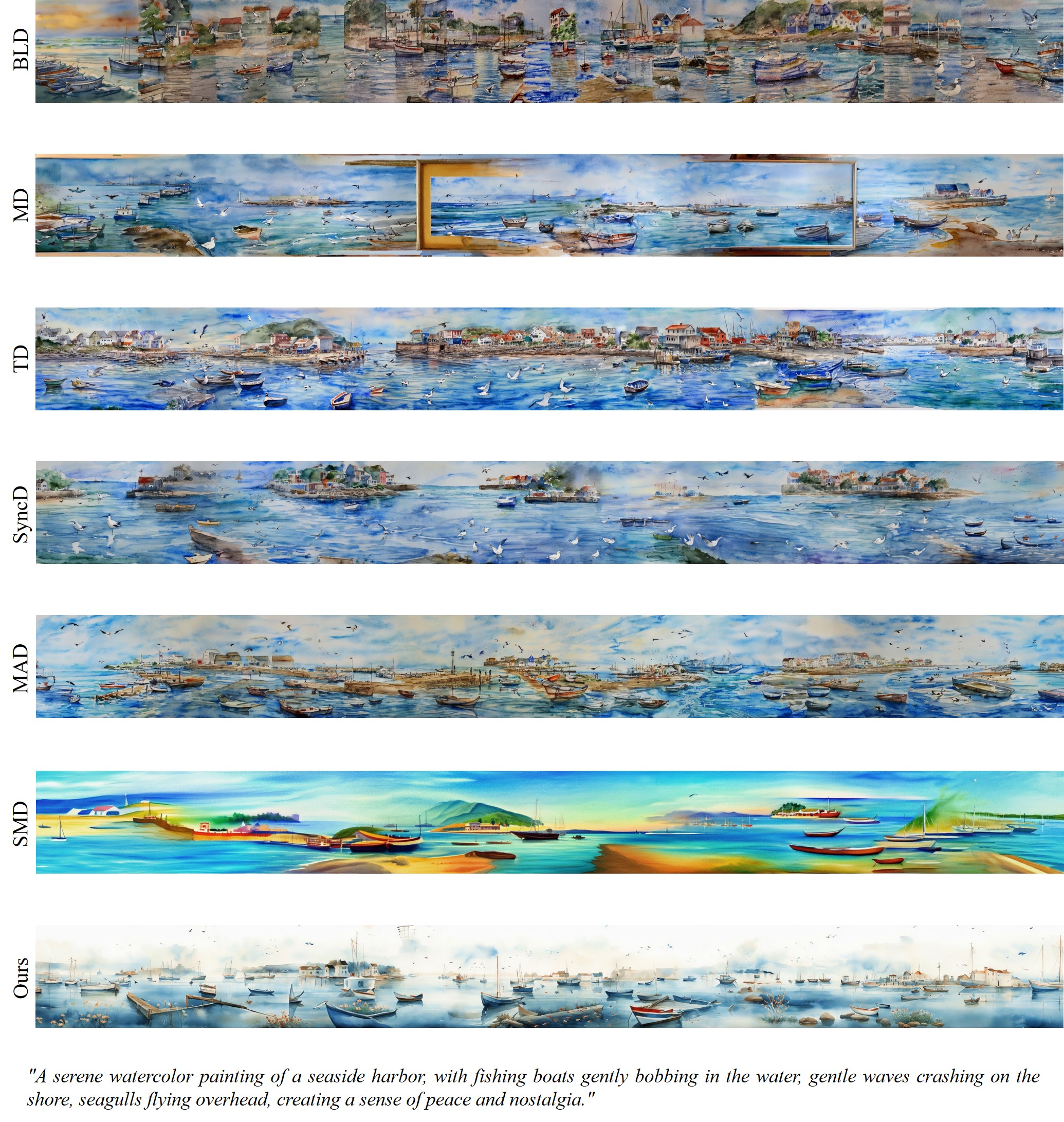}
    \caption{Additional qualitative comparison results on \(512\times5120\) panorama generation.}
    \label{fig:qual_comp_plus_2}
\end{figure}

\begin{figure}[!htbp]
    \centering
    \includegraphics[width=\linewidth]{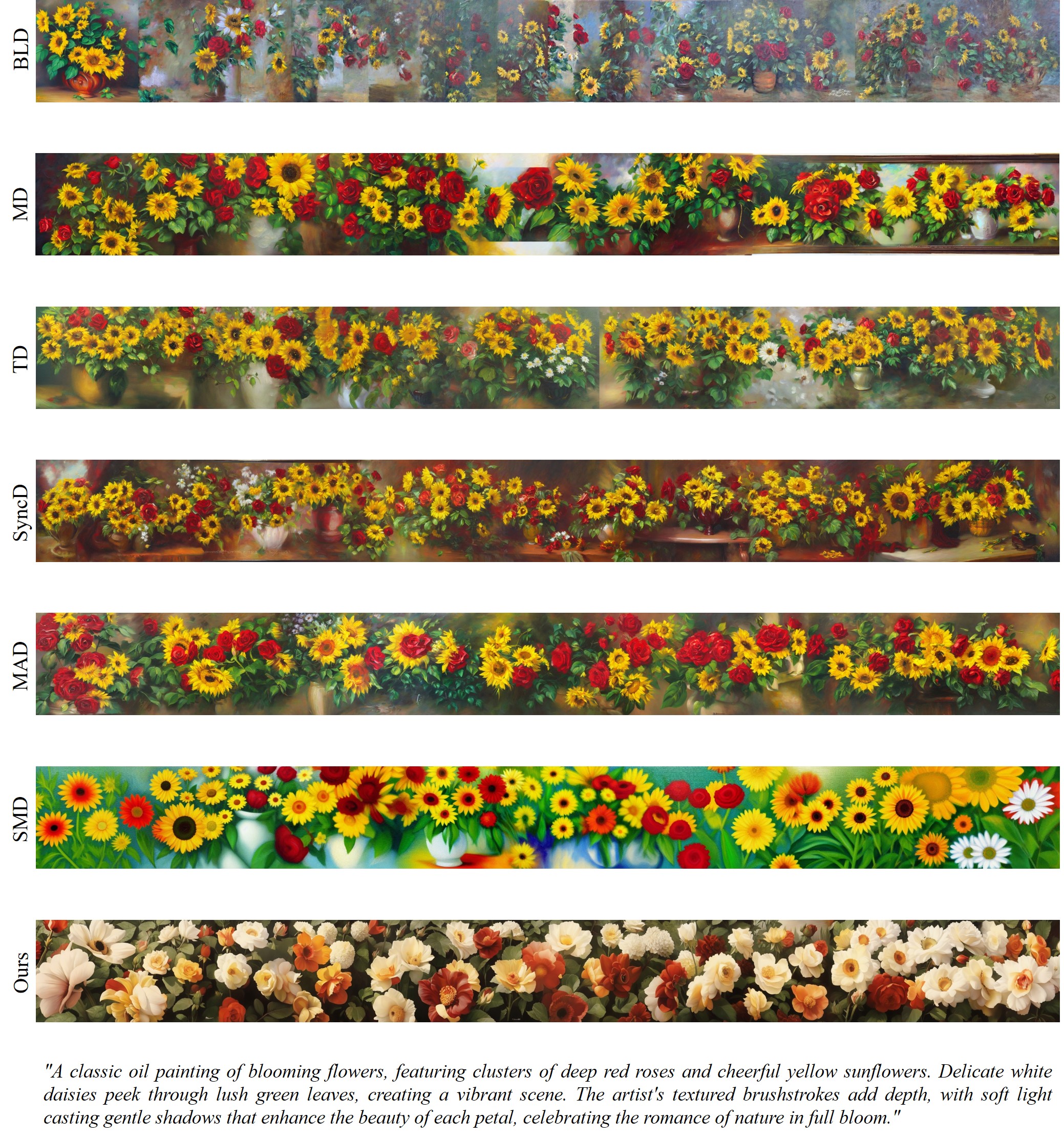}
    \caption{Additional qualitative comparison results on \(512\times5120\) panorama generation.}
    \label{fig:qual_comp_plus_3}
\end{figure}

\begin{figure}[!htbp]
    \centering
    \includegraphics[width=\linewidth]{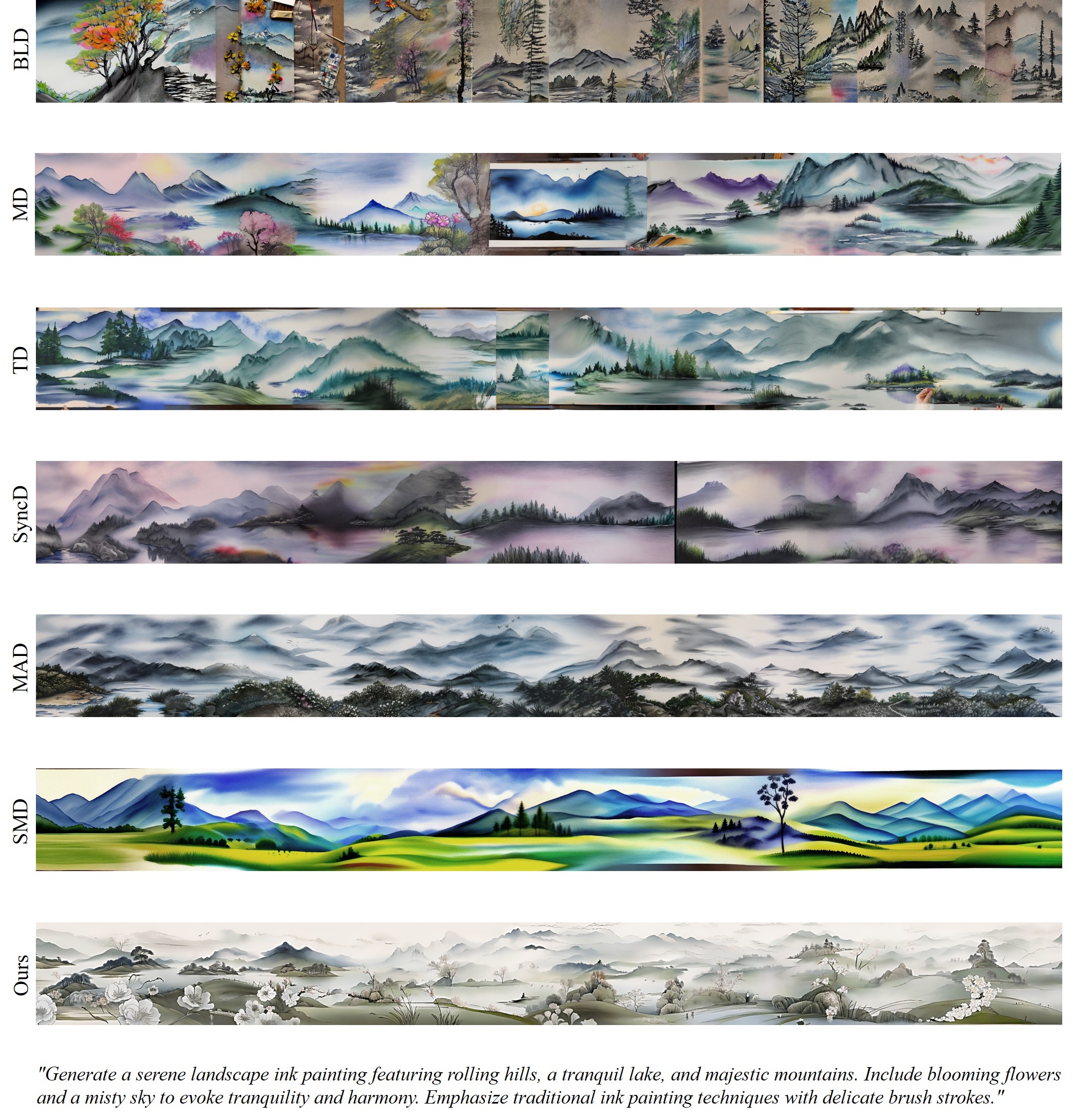}
    \caption{Additional qualitative comparison results on \(512\times5120\) panorama generation.}
    \label{fig:qual_comp_plus_4}
\end{figure}

\begin{figure}[!htbp]
    \centering
    \includegraphics[width=\linewidth]{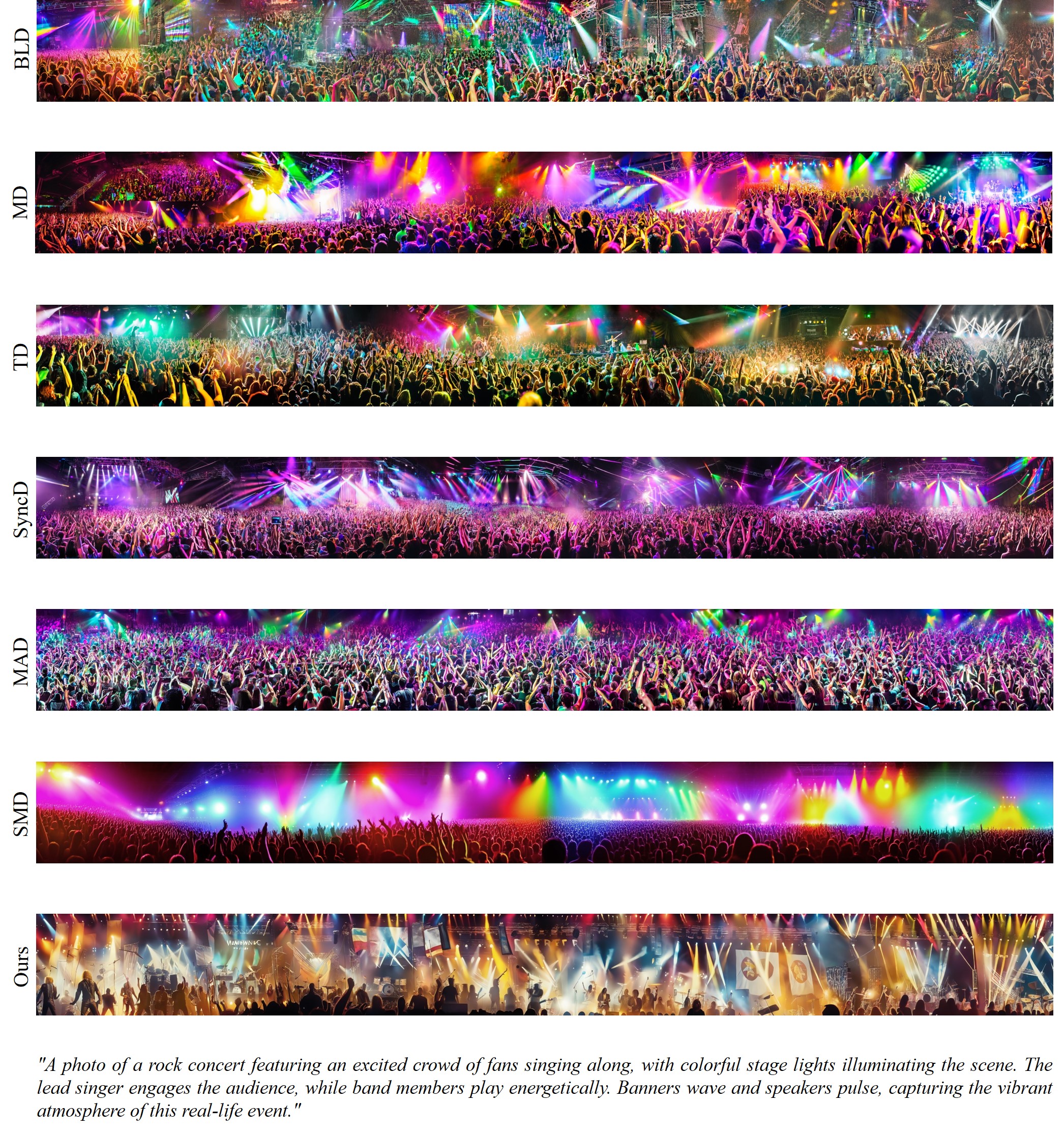}
    \caption{Additional qualitative comparison results on \(512\times5120\) panorama generation.}
    \label{fig:qual_comp_plus_5}
\end{figure}

\begin{figure}[!htbp]
    \centering
    \includegraphics[width=\linewidth]{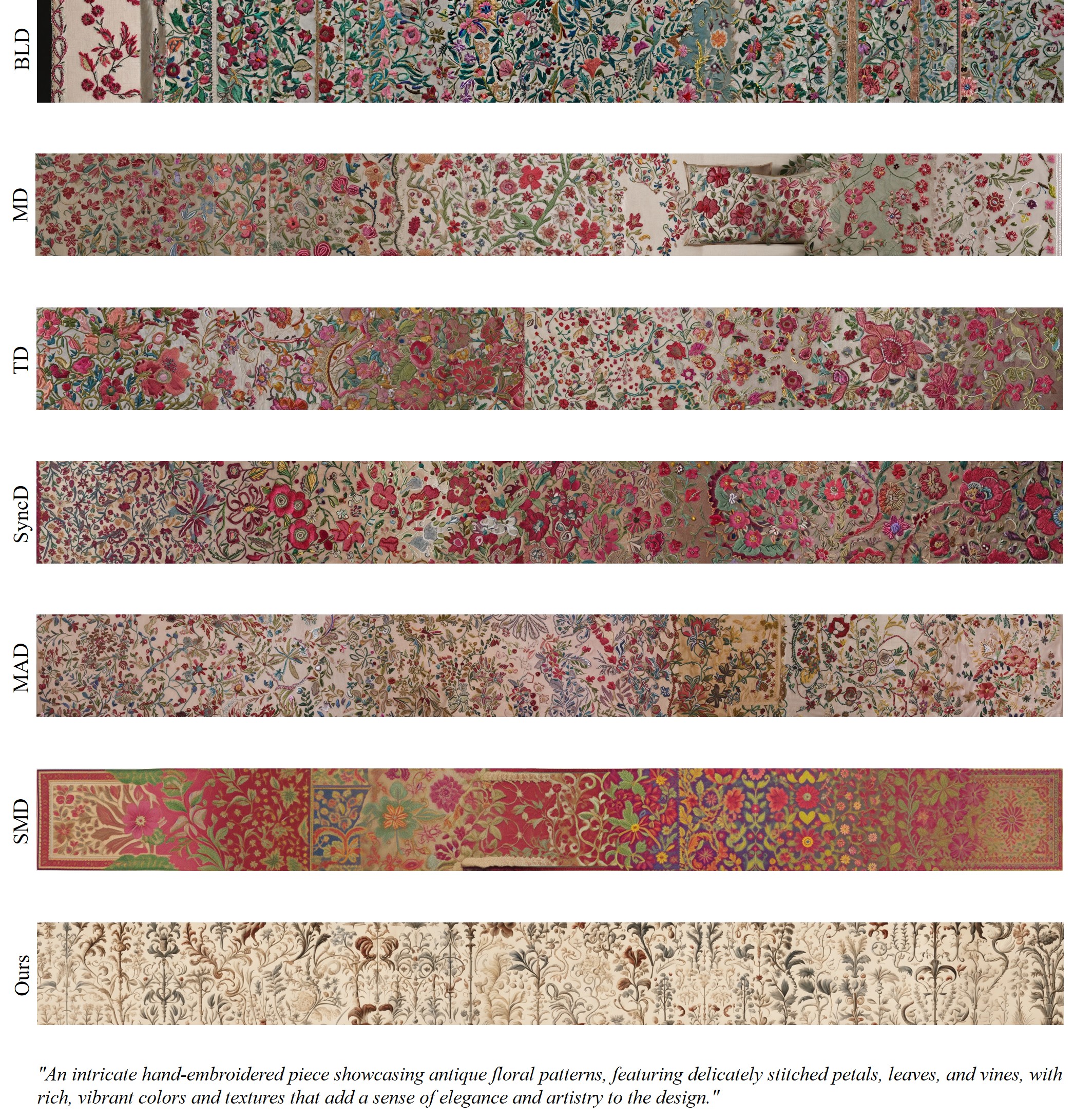}
    \caption{Additional qualitative comparison results on \(512\times5120\) panorama generation.}
    \label{fig:qual_comp_plus_6}
\end{figure}

\begin{figure}[!htbp]
    \centering
    \includegraphics[width=\linewidth]{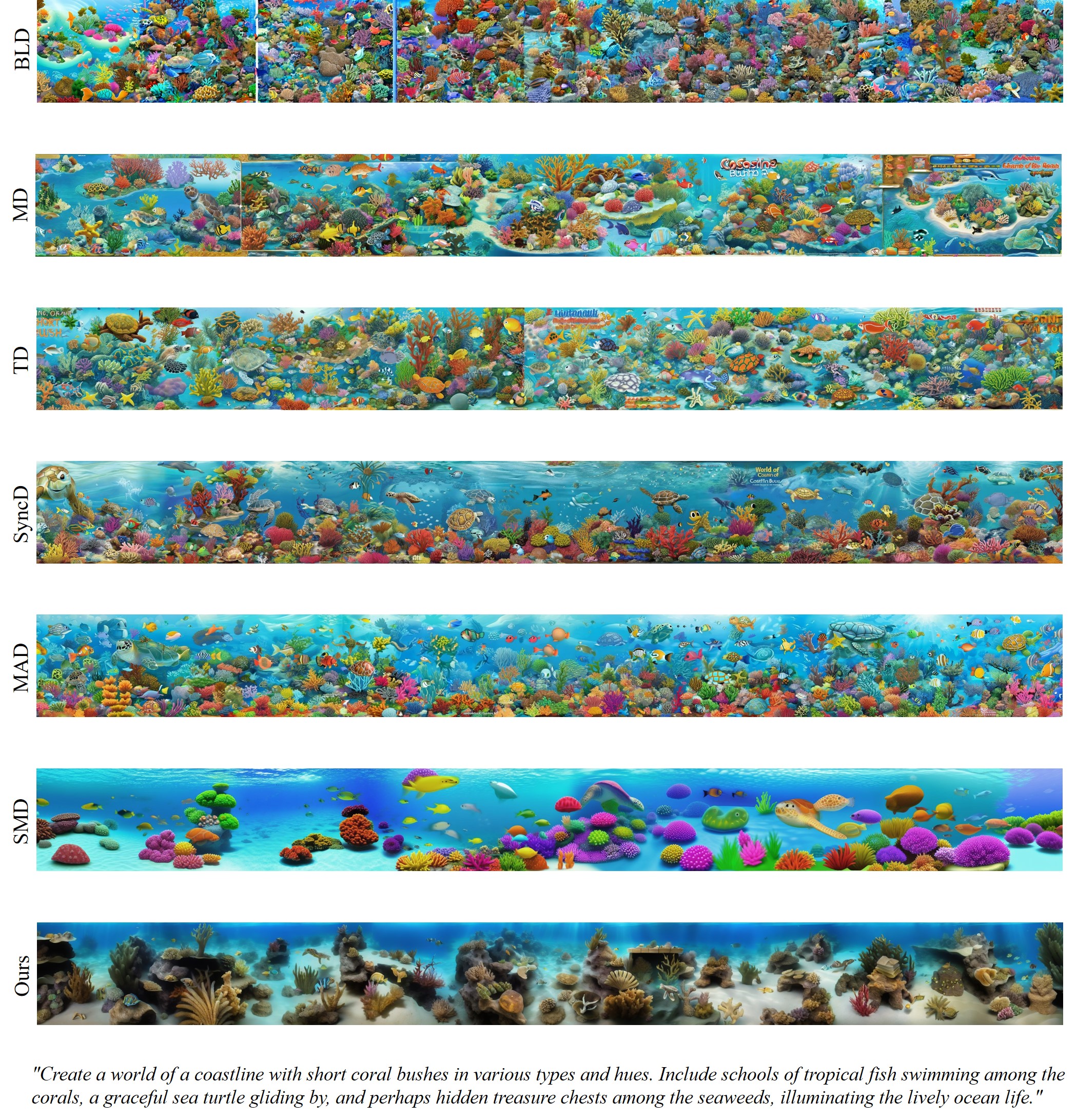}
    \caption{Additional qualitative comparison results on \(512\times5120\) panorama generation.}
    \label{fig:qual_comp_plus_7}
\end{figure}

\begin{figure}[!htbp]
    \centering
    \includegraphics[width=\linewidth]{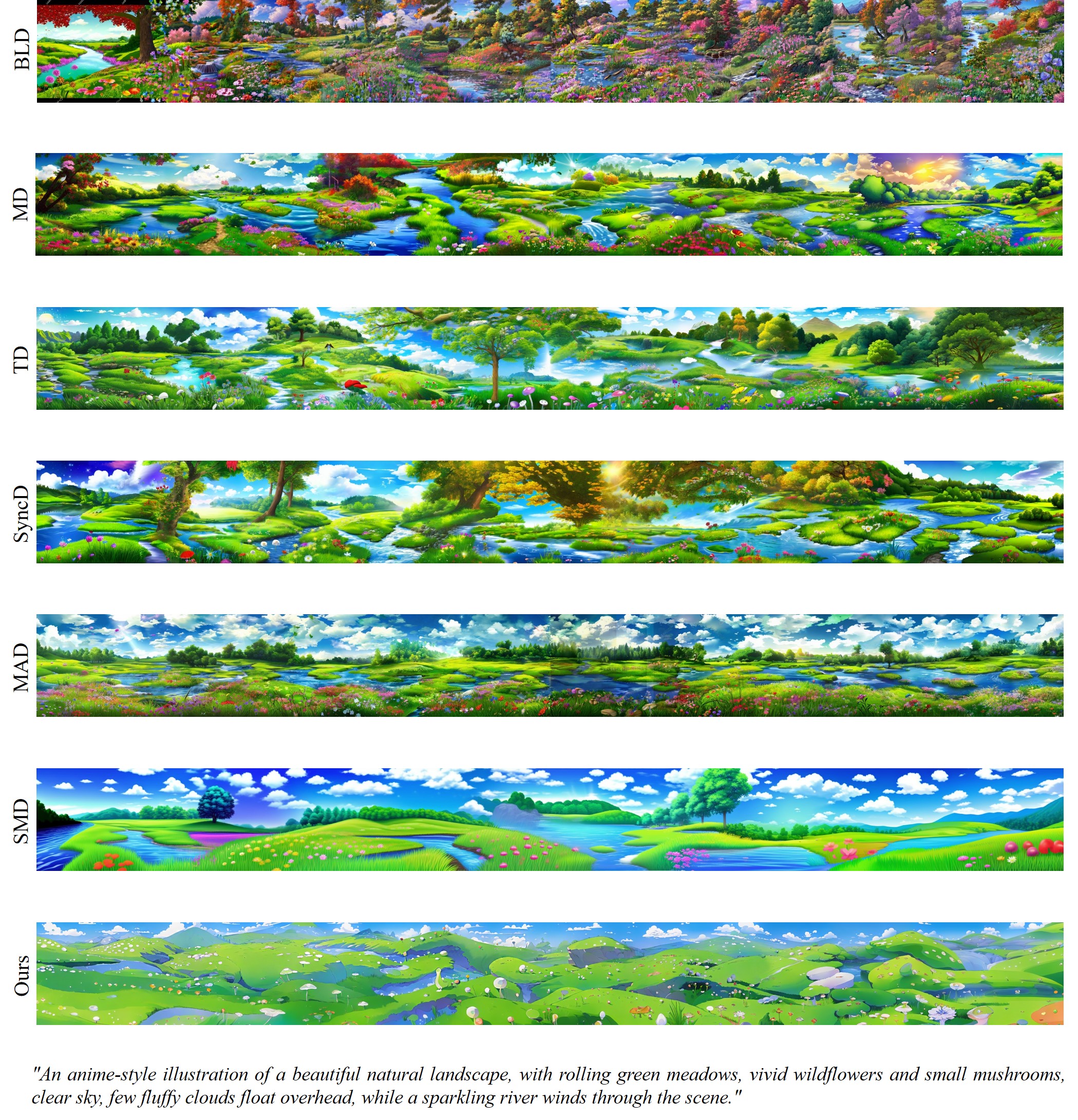}
    \caption{Additional qualitative comparison results on \(512\times5120\) panorama generation.}
    \label{fig:qual_comp_plus_8}
\end{figure}

\begin{figure}[!htbp]
    \centering
    \includegraphics[width=\linewidth]{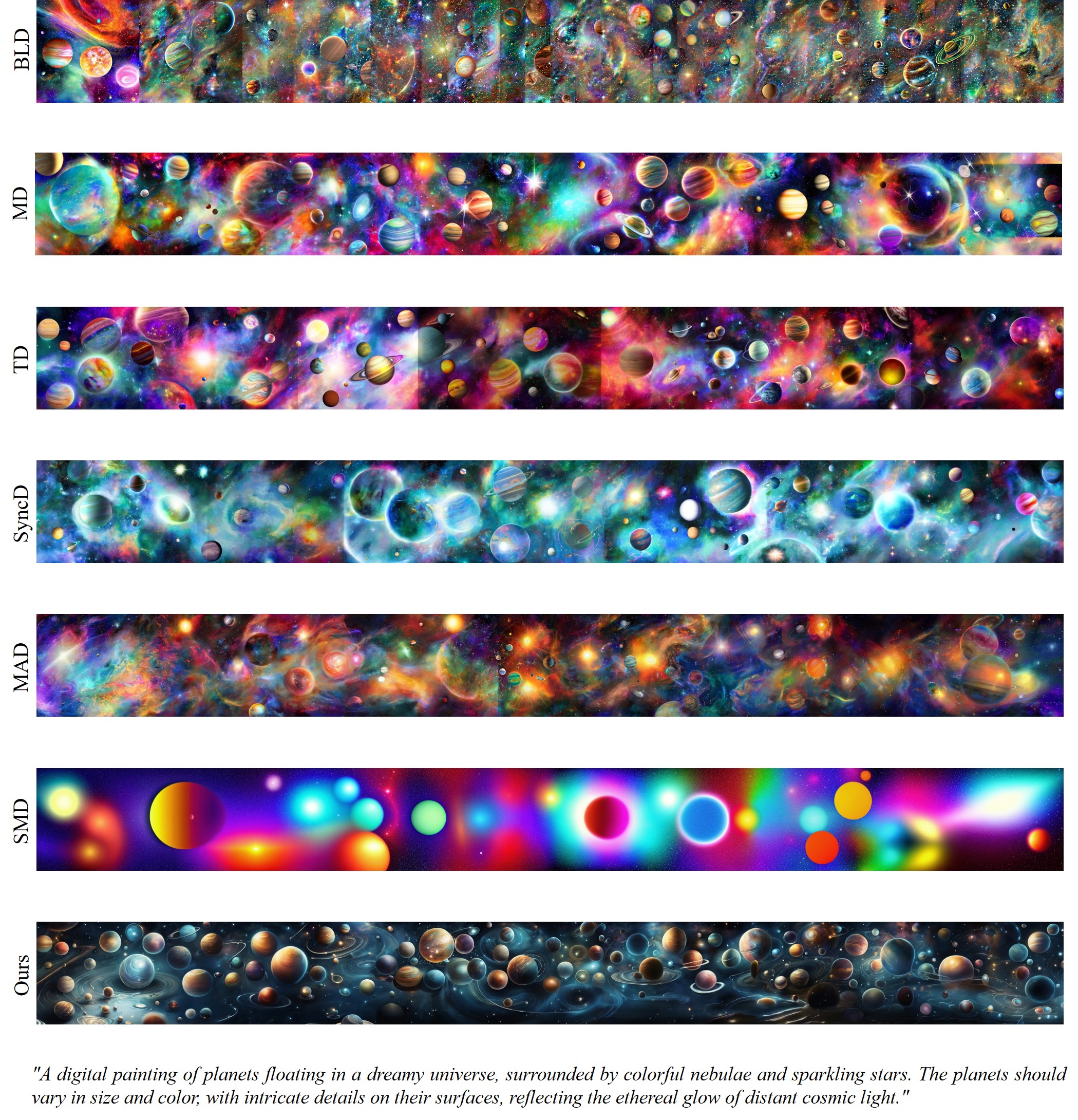}
    \caption{Additional qualitative comparison results on \(512\times5120\) panorama generation.}
    \label{fig:qual_comp_plus_9}
\end{figure}

\begin{figure}[!htbp]
    \centering
    \includegraphics[width=\linewidth]{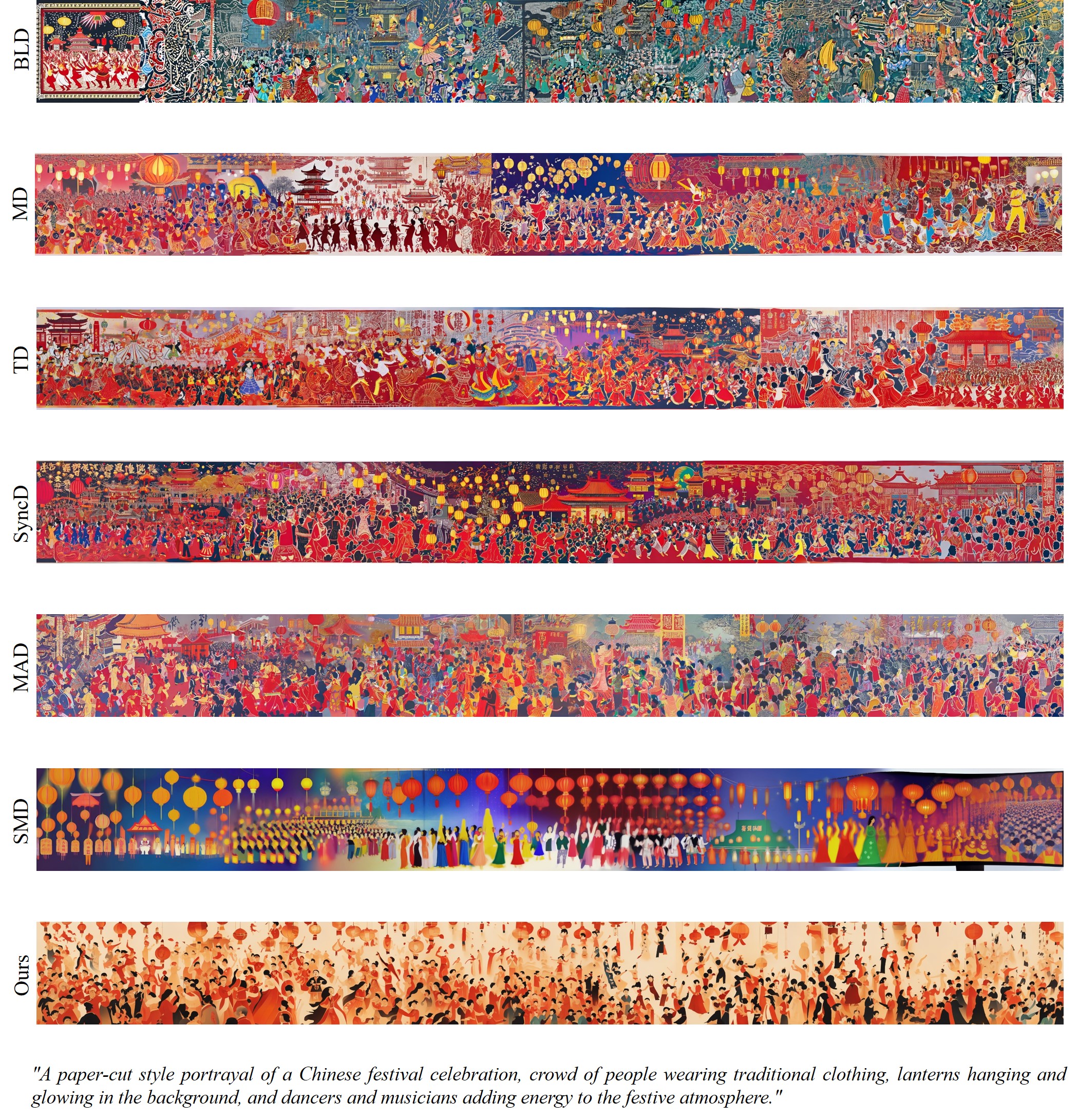}
    \caption{Additional qualitative comparison results on \(512\times5120\) panorama generation.}
    \label{fig:qual_comp_plus_10}
\end{figure}

\begin{figure}[!htbp]
    \centering
    \includegraphics[width=\linewidth]{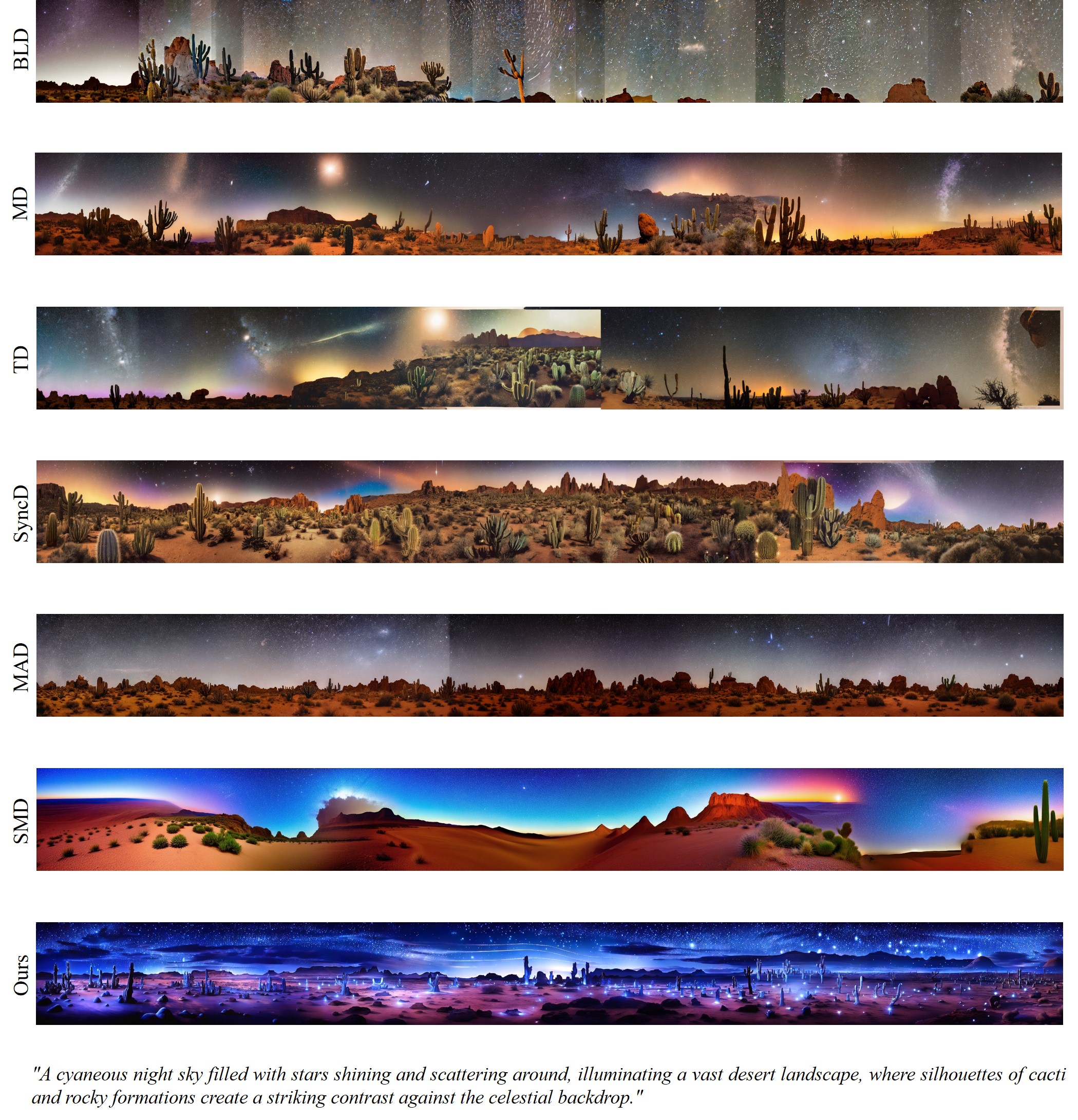}
    \caption{Additional qualitative comparison results on \(512\times5120\) panorama generation.}
    \label{fig:qual_comp_plus_11}
\end{figure}

\begin{figure}[!htbp]
    \centering
    \includegraphics[width=\linewidth]{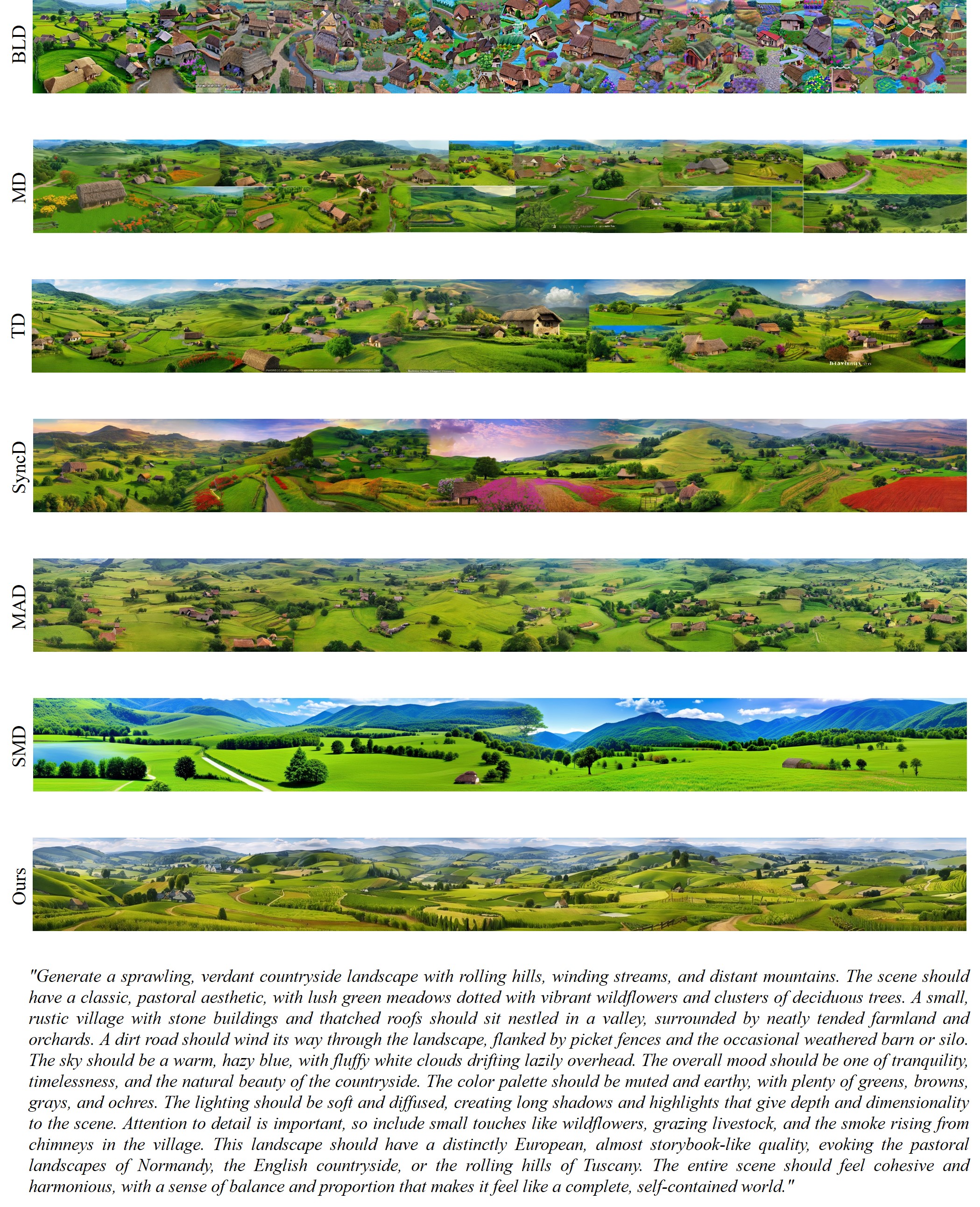}
    \caption{Additional qualitative comparison results on \(512\times5120\) panorama generation.}
    \label{fig:qual_comp_plus_12}
\end{figure}

\cref{fig:qual_comp_plus_0,fig:qual_comp_plus_1,fig:qual_comp_plus_2,fig:qual_comp_plus_3,fig:qual_comp_plus_4,fig:qual_comp_plus_5,fig:qual_comp_plus_6,fig:qual_comp_plus_7,fig:qual_comp_plus_8,fig:qual_comp_plus_9,fig:qual_comp_plus_10,fig:qual_comp_plus_11,fig:qual_comp_plus_12} gives more qualitative comparisons on panoramic image generation of \(512\times5120\), highlighting the areas where improvements have been made.

\vspace*{\fill}
\vfill
\vspace*{\fill}
\vfill
\vspace*{\fill}
\vfill
\vspace*{\fill}
\vfill
\vspace*{\fill}
\vfill
\vspace*{\fill}
\vfill
\vspace*{\fill}
\vfill
\vspace*{\fill}
\vfill
\vspace*{\fill}
\vfill
\vspace*{\fill}
\vfill
\vspace*{\fill}
\vfill
\vspace*{\fill}
\vfill
\vspace*{\fill}
\vfill

\clearpage
\section{Comparison with SD Variants}
\label{appendix:phi_abl}


We further conduct experiments on additional diffusion variants, evaluating baselines using \(\Phi=\text{SDXL}\). In line with \cref{sec:comparison}, we continue to generate images at a resolution of \(512\times5120\) and then crop them to create \(512\times512\) image sets.

\cref{tab:sdver_abl} presents the comparison results. Some baselines do not support SDXL and are therefore excluded from this table. Notably, even with SDXL, our approach excels at enhancing the multilevel coherence of generated panoramas, though to a lesser degree than the SD-based reference model. This is expected, given SDXL's advanced design as a diffusion model with doubled default resolution in both height and width.

\begin{table}[!htbp]
    \fontsize{11pt}{16pt}\selectfont
    \centering
    \caption{Comparisons between PanoLlama and baselines with another \(\Phi\) variants. Our approach still stands out in improving the coherence of the generated panoramas while maintaining aesthetic quality and operational speed.}
    \label{tab:sdver_abl}
    \begin{adjustbox}{width=\textwidth}
	\begin{tabular}{ccccccccc}
            \hline
            \multirow{2}{*}{} & \multicolumn{4}{c}{Coherence} & \multicolumn{2}{c}{Fidelity \& Diversity} & \multicolumn{2}{c}{Compatibility} \\
            \cmidrule(lr){2-5} \cmidrule(lr){6-7} \cmidrule(lr){8-9}
    	& LPIPS\(\downarrow\) & DISTS\(\downarrow\) & TV\(\downarrow\) & SSIM\(\uparrow\) & FID\(\downarrow\) & IS\(\uparrow\) & CLIP\(\uparrow\) & CLIP-aesthetic\(\uparrow\) \\
    	\hline
            SD\(_\text{XL}\) & -- & -- & -- & -- & 34.50 & 7.60 & 33.23 & 6.76 \\
            LlamaGen & -- & -- & -- & -- & 37.82 & 6.43 & 31.62 & 6.74 \\
            \hdashline
            BLD\(_\text{XL}\) & 0.790 & 0.356 & 0.075 & 0.013 & 85.20 (+50.70) & 5.97 (-1.63) & 32.59 (-0.64) & 5.80 (-0.96) \\
            MD\(_\text{XL}\) & 0.676 & 0.253 & 0.055 & 0.220 & \underline{39.15 (+4.65)} & 6.42 (-1.18) & \underline{34.66 (+1.43)} & 6.88 (+0.12) \\
            TD\(_\text{XL}\) & 0.638 & 0.214 & 0.051 & 0.274 & 40.05 (+5.55) & \underline{6.48 (-1.12)} & \textbf{34.79 (+1.56)} & 6.89 (+0.13) \\
    	MAD\(_\text{XL}\) & \underline{0.517} & \underline{0.208} & \underline{0.032} & \underline{0.296} & 56.55 (+22.05) & 5.13 (-2.47) & 32.07 (-1.16) & \underline{6.94 (+0.18)} \\
            Ours & \textbf{0.410} & \textbf{0.196} & \textbf{0.021} & \textbf{0.305} & \textbf{40.09 (+2.27)} & \textbf{5.97 (-0.46)} & 31.56 (-0.06) & \textbf{6.97 (+0.23)} \vspace{0.04cm} \\
    	\hline
	\end{tabular}
 \end{adjustbox}
\end{table}

\section{Dataset Construction Details}
\label{appendix:dataset}

To ensure a comprehensive and fair evaluation, our dataset construction focused on diversity and challenge. (i) Theme Selection: We select 25 common themes stratified by typical scene scale: from expansive scenes well-suited for panoramas (e.g., 'landscape') to medium-scale subjects (e.g., 'architecture') and challenging dense/small-object scenes (e.g., 'crowd'). This variety is designed to robustly test PIG methods across diverse content types, as explored in our prompt theme analysis in \cref{sec:ablation}. (ii) Sub-Themes: Each theme is further diversified into 3-8 sub-themes (e.g., 'crowd' includes 'festival', 'market', 'concert', \(\cdots\)) to ensure broad conceptual coverage. (iii) Styles: We include a mix of styles (photorealistic, artistic, chosen randomly). (iv) Creation Process: The prompts are generated by an AI based on the aforementioned structured themes and styles, resulting in 400 prompts per theme. (v) Fairness: This random, broad design ensures no implicit bias toward our PanoLlama; all methods are evaluated fairly.

\section{Implementation Details about Our Applications}
\label{appendix:applications}


Similar to joint diffusion methods, our approach can also introduce additional optimizations on top of next-token prediction to achieve smoother transitions. For instance, we can apply a basic blending function as follows:
\begin{equation}\label{eq:blending_token}
\begin{aligned}
    \bar e_{i,1}&=\lambda e_{i,1}+(1-\lambda)e_{i-1,p} \\
    v_{i,1}^*&=\arg\min_{v_{i,1}}\|\bar e_{i,1}-e\|^2
\end{aligned}
\end{equation}
where \(v_{i-1,p}\) and \(v_{i,1}\) represent the boundary tokens between \(v_{i-1}\) and \(v_i\), \(e_{i-1,p}\) and \(e_{i,1}\) refer to the embeddings indexed by these tokens, \(\lambda\) denotes the transition factor, \(e\) represents the trained embeddings in \(f_{\mathcal{T}}\), \(v_{i,1}^*\) is the resulting token after blending.

By employing \cref{eq:blending_token} with \(\lambda\in[0.5,0.8]\), we can achieve smoother transitions under different textual conditions for multi-layout and multi-guidance applications. However, our experiments demonstrate that with a consistent prompt, the performance of PanoLlama peaks at \(\lambda=1.0\), declining at other values. This suggests that our method reaches its best in single-prompt scenarios without the need for blending operations.




\end{document}